\documentclass[11pt]{article}

\usepackage[final]{acl}

\usepackage{times}
\usepackage{latexsym}

\usepackage[T1]{fontenc}
\usepackage[utf8]{inputenc}

\usepackage{microtype}

\usepackage{inconsolata}

\usepackage{graphicx}
\usepackage{microtype}
\usepackage{hyperref,amsmath}
\usepackage{url}
\usepackage{booktabs}

\usepackage{lineno}
\usepackage{graphicx} % Required for inserting images
\usepackage{algorithm}
\usepackage{algpseudocode}
\usepackage{amsmath}

\usepackage{graphicx}
\usepackage{amssymb} % Nécessaire pour \mathbb
\usepackage{amsmath}
\usepackage{linguex}

\usepackage{tikz}
\usepackage{xcolor}
\usepackage{amsmath,amsfonts,bm}

\def\eqref#1{equation~\ref{#1}}
\def\1{\bm{1}}

\DeclareMathAlphabet{\mathsfit}{\encodingdefault}{\sfdefault}{m}{sl}
\SetMathAlphabet{\mathsfit}{bold}{\encodingdefault}{\sfdefault}{bx}{n}

\newcommand{\softmax}{\mathrm{softmax}}

\DeclareMathOperator*{\argmax}{arg\,max}
\DeclareMathOperator*{\argmin}{arg\,min}

\usepackage{hyperref}
\newcommand{\hidden}[1]{}
\usepackage{url}
\usepackage{graphicx}
\usepackage[most]{tcolorbox}

\newtheorem{theorem}{Theorem}

\newtheorem{proposition}{Proposition}

\definecolor{jerorange}{RGB}{204,102,0} % un orange sombre inspiré des amplis Orange

\newcommand{\sgn}{{\rm sgn}}

\definecolor{darkblue}{rgb}{0, 0, 0.5}
\hypersetup{colorlinks=true, citecolor=darkblue, linkcolor=darkblue, urlcolor=darkblue}

\title{

SSA: Improving Performance With a Better Scoring Function}

\hidden{
\author{
\parbox{0.24\textwidth}{\centering
Omar Naim \\
Université de Toulouse \\
IRIT \\
\texttt{omar.naim.docs@gmail.com}
}
\hfill
\parbox{0.24\textwidth}{\centering
Swarnadeep Bhar \\
Université de Toulouse \\
IRIT \\
\texttt{swarnadeep.bhar@irit.fr}
}
\hfill
\parbox{0.24\textwidth}{\centering
Nicholas Asher \\
IRIT \\
\texttt{nicholas.asher@irit.fr}
}
\hfill
\parbox{0.24\textwidth}{\centering
Jérôme Bolte \\
Toulouse School of Economics \\
University of Toulouse Capitole \\
\texttt{jbolte@ut-capitole.fr}
}
}
}
\author{
Omar Naim \\
IRIT \\
{\small Université de Toulouse} \\
{\small \texttt{omar.naim.docs@gmail.com}}
\And
Swarnadeep Bhar \\
IRIT \\
{\small Université de Toulouse} \\
{\small \texttt{swarnadeep.bhar@irit.fr}}
\And
Jérôme Bolte \\
Toulouse School of Economics \\
{\small University of Toulouse Capitole} \\
{\small \texttt{jbolte@ut-capitole.fr}}
\And
Nicholas Asher \\
IRIT \\
{\small CNRS} \\
{\small \texttt{nicholas.asher@irit.fr}}
}
\begin{document}
\maketitle
\begin{abstract}

While transformer models exhibit strong in-context learning (ICL) abilities, they often fail to generalize under simple distribution shifts. We analyze these failures and identify Softmax, the scoring function in the attention mechanism, as a contributing factor. We propose \textbf{Scaled Signed Averaging (SSA)}, a novel attention scoring function that mitigates these failures. SSA significantly improves performance on our ICL tasks and outperforms transformer models with Softmax on several NLP benchmarks and linguistic probing tasks, in both decoder-only and encoder-only architectures.

\hidden{\color{orange}
While Large Language models' abilities for in-context learning (ICL) have had much success, they have limitations on simple semantic tasks involving quantifiers like {\em every} and {\em some}, as well as on tasks with linear functions. We analyze those limitations and identify Softmax, the scoring function in the attention mechanism, as a contributing factor to these limitations.  Our \textbf{scaled signed averaging (SSA)}, a novel scoring function mitigates these limitations.  SSA significantly improves performance on our ICL tasks. In addition, SSA outperforms transformer models with Softmax 
on several early learning NLP benchmarks and linguistic probing tasks.}

\end{abstract}

\section{Introduction}
Scoring functions are a core component of  the attention mechanism of transformer architectures, governing how information is aggregated across tokens and crucial for in-context %. Prior work has shown that attention is both necessary and sufficient for simple in-context 
learning (ICL).  % \cite{olsson:etal:2022,naim:asher:2024b}, making the choice of scoring function central to understanding how Transformers learn and generalize, both during training and at inference time.
Despite the empirical success of ICL, however, recent studies have identified systematic limitations in its generalization behavior \cite{mccoy:etal:2024,ye:etal:2024, analyzinglimits}. Large language models tend to perform well in contexts that resemble patterns frequently encountered during training, but struggle in so-called \emph{low-probability} settings---tasks, data distributions, or algorithmic structures that are rare or absent from the training corpus. %These failures point to fundamental constraints on the ability of ICL to extrapolate to genuinely novel situations.

We investigate the origins of these limitations in a controlled setting. We study two simple ICL tasks using transformer models trained from scratch, allowing us to isolate architectural effects from pretraining artifacts. We show that the use of the Softmax scoring function in attention is one contributor to the observed generalization failures.

To improve generalization, we introduce \emph{Scaled Signed Averaging} (SSA), a trainable alternative to Softmax for attention scoring. We show that SSA substantially improves ICL generalization in small transformer models in our controlled tasks. We further demonstrate that these gains extend beyond synthetic settings: a decoder-only Transformer model (Nemotron-style, GPT-2–like) trained from scratch on the \textit{FineWeb} corpus \cite{fineweb} with SSA achieves lower perplexity and stronger performance than a Softmax-based counterpart across several standard NLP benchmarks.
%We further demonstrate that these gains extend beyond synthetic settings: a decoder-only GPT-2--style model trained from scratch on the \textit{FineWebText} corpus \cite{fineweb} with SSA achieves lower perplexity and stronger performance than a Softmax-based counterpart across several standard NLP benchmarks for early-stage Transformer evaluation.
%, under both zero-shot and few-shot conditions. %All models are trained from scratch on the \textit{FineWebText} corpus \cite{fineweb}.
We also evaluate SSA in encoder architectures. In particular, we train several variants of BabyBERTa on a corpus of child-directed speech, including a standard Softmax model and SSA-based counterparts. Consistent with our decoder-only results, SSA-based BabyBERTa models achieve improved performance on multiple tasks in the grammatical probing suite of \cite{huebner:etal:2021}.

\section{Related Work}

\paragraph{In-Context-Learning}

\citet{brown:etal:2020} introduced ICL as a paradigm in which a model learns at inference time from the prompt by analogy, without modifying any training parameters. \citet{dong2022survey} survey the successes and challenges of ICL, noting that existing research has primarily focused ``on simple tasks and small models'', such as learning linear or basic Boolean functions. A reason for this focus is that ICL emerges through training, so studying it requires training from scratch. In this work, we investigate both types of tasks.

\paragraph{Function Learning and Quantification}
Transformers trained from scratch can perform ICL of simple functions under favorable conditions. \citet{garg:etal:2022} demonstrated successful ICL of linear functions when training and test distributions are matched, and \citet{bhattamishra2023understanding} showed ICL of Boolean functions in small GPT-2–style models. \citet{raventos2024pretraining} analyzed how ICL capabilities evolve with pretraining scale under matched distributions. For quantification, \citet{asher:etal:2023} proposed encoding semantic situations as input sequences to assess a generative model's understanding of basic quantifiers. We adopt both of these evaluation paradigms: we test linear function prediction under distribution shift, and probe a model's grasp of ``\textit{every}'' and ``\textit{some}'' by encoding quantified situations in context and evaluating the correctness of the model's interpretations.

\paragraph{Limits of In-Context-Learning}
 A growing body of work highlights sharp limits of ICL under distribution shift. Performance degrades substantially when inference-time inputs differ from the training distribution \citep{xie2021explanation, zhang:etal:2024, giannou:etal:2024}.  \citet{naim:asher:2024b} systematically characterized distribution shifts causing significant degradation, showing the failure is neither overfitting nor memorization.
\citet{ye:etal:2024, mccoy:etal:2024} demonstrate broader limits inherent to autoregressive training.  \citet{analyzinglimits} extend this analysis to provide theoretical characterizations of ICL failure modes. Our work complements these findings by identifying a concrete architectural mechanism, Softmax saturation, as a root cause, and by proposing a targeted fix.

\paragraph{Problems with Softmax}  
 
Prior work has documented concentration effects inherent to the Softmax operation in attention: weights often collapse onto a small set of co-occurring tokens, neglecting others, and in early layers many heads attend almost exclusively to the first token \citep{ref1,ref2,ref3}. Beyond token-level collapse, models may rely on only a few tokens, or even a single token per prompt, for prediction \citep{ref4}, with decision weight disproportionately concentrated on the final token \citep{ref5}. Several alternative normalization functions have been proposed to address these limitations. Sparsemax \citep{sparsemax} replaces Softmax with a projection onto the probability simplex, yielding sparse attention distributions that can zero out irrelevant tokens entirely. Entmax \citep{entmax} generalizes both Softmax and Sparsemax through a parametric family, enabling controllable sparsity between dense and sparse attention regimes. Temperature-scaled Softmax \citep{hinton2015distilling} adjusts the entropy of the attention distribution via a scalar temperature parameter. More recent approaches include SA-Softmax \citep{zheng:etal:2025}, which adapts logit scaling dynamically, and CosFormer \citep{qin:etal:2022}, which replaces the exponential kernel with a cosine-modulated linear alternative. Despite their diversity, we show in Section~\ref{sec:alternatives-main} that none of these alternatives yields consistent improvements over standard Softmax on our tasks, motivating the design of SSA.

\paragraph{NLP Benchmarks} \citet{huebner:etal:2021} demonstrate that transformer-based masked language models can effectively learn core grammatical structures from a small, child-directed corpus. their BabyBERTa achieves a grammatical understanding comparable to RoBERTa-base pre-trained on 30B words. \cite{huebner:etal:2021} also develop a grammar evaluation suite tailored to child-level vocabularies. We use this suite to compare SSA and Softmax as scoring functions in encoder-only models.

\section{Our ICL Tasks and Experimental Setup}

\subsection{Tasks}
\label{sec:tasks}
We introduce two controlled ICL tasks and experimental setup used throughout the paper. The aim is to isolate the role of the attention scoring function and determine whether generalization failures arise from architectural choices rather than data complexity or scale.

The tasks probe complementary capabilities: logical aggregation and functional inference. Both tasks use synthetic data, operate on sequences of numbers and use transparent prompting to minimize confounds from prompt design.

\paragraph{Quantification task.}

This task evaluates whether a model can correctly interpret simple quantifiers over a sequence (see Figure \ref{attn-map}). Given a sequence of numbers, the model must predict the truth of statements such as:

\ex. 
\a. \label{every} Every number in the sequence is positive. 
\b. \label{some} Some number in the sequence is positive. 

Training sequences $S$ are drawn from an input distribution $D_{\cal I} = \mathcal{N}(0,1)$. 

At test time, we evaluate generalization under two shifts: (i) longer sequences $S^{\text{test}}$ with lengths ranging from 10 to 200, and (ii) changes in input scale, with $D^{\text{test}}_{\cal I} = \mathcal{N}(0,\sigma)$ for $\sigma \geq 1$. 

%This setup tests whether the model can reliably aggregate information across longer contexts and varying input magnitudes.

\paragraph{Linear function task.}

In this task, the model must infer an affine function $f(x) = ax + b$ from in-context examples. Coefficients $(a,b)$ are sampled from $D_{\cal F} = \mathcal{N}(0,1)$, and inputs $x_i$ are drawn independently from $D_{\cal I}$.

At test time, we vary both the input distribution $D^{\text{test}}_{\cal I} \sim \mathcal{N}(0,\sigma_1)$ and the function distribution $D^{\text{test}}_{\cal F} \sim \mathcal{N}(0,\sigma_2)$, with $\sigma_1, \sigma_2 \geq 1$.

%This task isolates the ability to infer functional relationships from context and generalize beyond the training distribution.

\subsection{Why these tasks?}

Despite their simplicity, our tasks are conceptually fundamental. %We did not choose our two tasks by accident. Though simple, they are rather fundamental. 
Failure on the quantification task suggests deep limitations in a model’s reasoning capabilities. The notion of logical or semantic consequence, for instance, involves quantification; a failure to understand quantification implies a failure to understand what it means to reason correctly and will entail mistakes in tasks like question answering \citep{chaturvedi:etal:2022}. 
%This limitation can lead to both obvious and subtle errors in downstream tasks such as question answering \cite{chaturvedi:etal:2022}.
%Quantification lies at the core of logical and semantic inference; without grasping it, a model cannot reliably reason in a logically valid way. This limitation can lead to both obvious and subtle errors in downstream tasks such as question answering \cite{chaturvedi:etal:2022}.
%If models fail on the quantification task, their performance on reasoning tasks will most likely suffer in both obvious and unobvious ways. 

%if you don’t understand quantification you don’t understand what it means to reason in a logically correct fashion. 
%Thus failures on the quantification task have
%important downstream consequences for real world tasks.

The function prediction task tests a model’s
ability to extrapolate patterns from contextual data to novel situations, a core requirement of ICL.  While large models often succeed by relying on extensive encoded knowledge, true generalization requires the ability to go beyond the training distribution. Our function task isolates this challenge in a controlled setting. If a model fails to generalize here, we should be cautious about claims of generalization in more complex, less controlled scenarios involving noisy or unknown data.

%Together, they provide a minimal testbed for assessing whether attention mechanisms support robust generalization beyond the training distribution.

\subsection{Training and Evaluation Setup}

We train models from scratch on sequences of input-output pairs 
$(x_1, f(x_1), \dots, x_i)$ followed by a query input $x_i$, for which the model must predict $f(x_i)$. Sequence lengths are sampled using a curriculum ranging from 11 to 40.\footnote{The code is available at \url{https://github.com/omyokun/SSA/}.}

We use decoder-only transformer models (GPT-2 style), ranging from 1 to 18 layers. Unless otherwise specified, we report results for a 12-layer model with 8 attention heads ``12L8AH'' (22.5M parameters), with and without MLP components, as larger models did not yield qualitatively different results.

To identify which components are responsible for 
ICL and its limitations, we performed ablation studies by systematically removing architectural components.

Training minimizes the expected autoregressive loss: %\jer{remove min in the formula improves clarity}
%\begin{equation} \label{eq:autoregressive}
%\small
%\min_{\theta} \ 
\[
\underset{\substack{f \sim \mathcal{D}_f \\ x_1, \dots, x_n \sim \mathcal{D}_I}}{\mathbb{E}}
\left[
\sum_{i=1}^{n} \ell\big(f(x_i), {\cal L}_\theta(x_1, f(x_1), \dots, x_i)\big)
\right]
\]
%\end{equation}
where $\ell$ is squared error for the linear task and cross-entropy for the quantification task. Models are trained for 500{,}000 steps with batch size 64.

\paragraph{Evaluation.}
For the quantification task, we evaluate performance over pairs 
$(S^{\text{test}}, D^{\text{test}}_{\cal I})$ by generating 100 
samples (each consisting of 64 batches) and reporting the average 
error rate. For the linear function task, we sample $100$ 
functions from $D^{\text{test}}_{\cal F}$. For each function, we 
generate batches of input sequences drawn from 
$D^{\text{test}}_{\cal I}$ and evaluate predictions at each 
position $\text{k} > 2$ given the preceding context. We report mean 
squared error averaged across all prediction points, batches, and 
sampled functions.

This protocol systematically measures generalization under 
distribution shift across both tasks. The following sections use 
this setup to expose systematic generalization failures, analyze 
their origins, and ultimately identify the architectural component 
responsible.

%\section{Our ICL tasks and experimental set up}

%f length $n$ and they first attempt to solve the task on the initial segment $(x_1)$, then on $(x_1, x_2)$ all the way to $(x_1... x_n)$.  We trained several transformers of different layer sizes (L) (from 1 to 18) and attention heads (AH) (from 1 to 8) from scratch on our different tasks. The standard architecture we used was (12L8AH): 12 layers, 8 attention heads with an embedding size of 256.  Cross-entropy was used as the loss function for our NLP tasks: \ref{every} and \ref{some}, while least squares was used for linear functions. More details can be found in Appendix ...}
% We trained several small decoder only transformer models from scratch to perform ICL of linear functions.

%Our models feature from 1 to 18 layers (L) and from 1 to 8 attention heads (AH) with an embedding size of 64 to 256. 

%The model's task is to predict the next value for $f(x_i)$ through a prompt of type $(x_1,f(x_1),...,x_i)$. We refer to that prediction as $\fh(x_i)$.  

% also studied by \cite{garg:etal:2022} but also chose Gaussian test distributions that had different parameters.  We also looked at . 

%% ─────────────────────────────────────────────
\section{ICL Results}\label{sec:icl-results}
%% ─────────────────────────────────────────────

In this section, we use the controlled setup introduced in Section \ref{sec:tasks} to evaluate whether models generalize under distribution shift across both tasks.

\vspace{0.5em}
\noindent\textbf{ICL with quantifiers.} When test samples are drawn from 
the same distribution as training, i.e.\ $D_{\cal I}, D^{\text{test}}_{\cal I} 
\sim {\mathcal N}(0,1)$, models successfully predict the correct truth 
values for \ref{every} and \ref{some}, even for test sequences 
$S^{\text{test}}$ substantially longer than those seen during training 
(Figure~\ref{hmap1}). However, performance drops sharply when inputs 
include one or more $x_i$ values far outside the training distribution 
(Figure~\ref{attn-map}). We refer to such sequences as \emph{deviant}.

\vspace{0.5em}
\noindent\textbf{ICL with linear functions.} We replicate the findings 
of \cite{naim:asher:2024b}: when training and test data are both sampled 
from ${\mathcal N}(0,1)$, even small models achieve near-zero average 
error. All models exhibit systematic non-zero errors when the target 
function is drawn from a shifted distribution $D^{\text{test}}_{\cal F} 
= {\mathcal N}(0, \sigma)$ with $\sigma > 2$ 
(Appendix~\ref{sec:appendixC}).

\vspace{0.5em}
\noindent\textbf{Takeaway.} Across both tasks, models perform well 
in-distribution but fail systematically under distribution shift, despite 
the underlying rule remaining unchanged. The following section analyzes 
the source of these failures.

%% ─────────────────────────────────────────────
\section{Error Analysis}\label{sec:error}
%% ─────────────────────────────────────────────

Having seen failures under distribution shift, we identify in this section what goes wrong when models 
encounter deviant sequences, and locate the failure within the 
architecture.

\vspace{0.5em}
\noindent\textbf{What goes wrong.} Across both tasks, the failure pattern 
is the same: a single sufficiently large value dominates the model's 
output, causing it to ignore other elements crucial for correct 
prediction. In the quantification task, models base their predictions 
for an entire deviant sequence $S$ almost exclusively on the largest 
element in $S$. The presence of a single such number is enough to 
trigger this behavior consistently (Figure \ref{attn-map}). In the linear function task, a 
single out-of-range input value similarly disrupts predictions across 
the entire sequence (Figure~\ref{sequence} and Table~\ref{table:LF1}).

\vspace{0.5em}
\noindent\textbf{Attention is the locus of failure.} Crucially, this 
behavior appears in both attention-only and full transformer models 
across all our training and testing setups. As with \cite{olsson:etal:2022}, 
ICL was effective even in models composed solely of attention layers, 
with no feedforward components (FF); these attention-only models performed 
comparably to their full transformer counterparts 
(Figure~\ref{progressive-lossAL}). In contrast, models consisting only 
of FF layers failed to perform ICL entirely. 
The attention mechanism is 
therefore both necessary and sufficient for ICL on our tasks. Since the 
same failure pattern appears with and without FF components, it cannot 
be attributed to the MLP but originates in the attention mechanism 
itself.

To rule out further representational issues, we verified that models 
could correctly classify individual numbers in deviant sequences as 
positive or negative (Figure~\ref{fig:sign-classification}). They performed well on this subtask, confirming 
that the information needed for correct prediction was available, but 
could not be used in the right way.

\hidden{
We also used a linear embedding 
$\mathrm{emb}: x \mapsto x \cdot W$, for $(x,W) \in \mathbb{R} \times 
\mathbb{R}^d$, which preserves magnitude ordering ($|x| < |y| \Rightarrow 
\|\mathrm{emb}(x)\| < \|\mathrm{emb}(y)\|$) and introduces no boundary 
effects, confirming that the failure must originate elsewhere in the 
architecture.
}
\vspace{0.5em}
\noindent\textbf{The failure is not a matter of scale.} We observed 
similar issues with larger pre-trained models. We evaluated performance 
on the quantification task using both fine-tuned and prompted versions 
of LLaMA 3.1 8B, as well as the prompted LLaMA 3.3 70B 
model.\footnote{Prompts are provided in Appendix~\ref{promptsAPP}.} 
In a 5-shot setting, prompted LLaMA 3.1 8B failed to master numerical 
inputs from $D^{\text{test}}_{\cal I}$ and showed no generalization to 
longer sequences. LLaMA 3.3 70B performed better on numerical inputs 
drawn from distributions outside ${\mathcal N}(0,1)$ but similarly 
failed to generalize to longer sequence lengths. Interestingly, the 
fine-tuned LLaMA 3.1 8B handled large numbers within a sequence 
(Figure~\ref{hmap3}), but still did not generalize beyond sequence 
lengths seen in training. On the linear function task, prompted LLaMA 
3.3 70B sometimes appeared to apply a regression-like strategy but 
still underperformed relative to our small models 
(Table~\ref{table:LF}).

\vspace{0.5em}
\noindent\textbf{Takeaway.} The generalization failure is architectural, 
not a consequence of model scale or data representation. Since the 
attention mechanism is both necessary and sufficient for ICL, and the 
failure persists regardless of scale, the scoring function within 
attention becomes the prime suspect. We examine this in the next section.

\begin{figure*}[h]
    \centering
    \begin{tikzpicture}
        % Define a blue-to-yellow heatmap color scale (viridis-like)
        \definecolor{heat0}{RGB}{86,18,103}     % Deep Blue (Low)
        \definecolor{heat1}{RGB}{84,85,152}  % Blue-Purple
        \definecolor{heat2}{RGB}{240,231,31} % Bright Yellow (High)
        % Spacing value
        \def\spacing{1.5} % Adjust this for more spacing

        \node[anchor=east] at (-0.8,1) {\textbf{Input:}};
        \node[ text=black, inner sep=8pt, minimum width=40pt, minimum height=25pt] at (1.5*\spacing,1) {(1,True,-2, False, 3, False, \textbf{70})};
        % Adjusted position of the Input label
        \node[anchor=east] at (-0.8,0) {\textbf{Attention weights:}};    
        % Third row: Attention Weights (now with new values)
        \node[fill=heat0, text=white, inner sep=8pt, minimum width=40pt, minimum height=25pt] at (0.1*\spacing,0) {1};
        \node[fill=heat0, text=white, inner sep=8pt, minimum width=50pt, minimum height=25pt] at (1.2*\spacing,0) {True};
        \node[fill=heat1, text=white, inner sep=8pt, minimum width=40pt, minimum height=25pt] at (2.3*\spacing,0) {-2};
        \node[fill=heat1, text=white, inner sep=8pt, minimum width=50pt, minimum height=25pt] at (3.4*\spacing,0) {False};
        \node[fill=heat0, text=white, inner sep=8pt, minimum width=40pt, minimum height=25pt] at (4.5*\spacing,0) {3};
        \node[fill=heat0, text=white, inner sep=8pt, minimum width=50pt, minimum height=25pt] at (5.6*\spacing,0) {False};
        \node[fill=heat2, text=black, inner sep=8pt, minimum width=50pt, minimum height=25pt] at (6.85*\spacing,0) {\textbf{70}};
        \node[anchor=east] at (-0.8,-1) {\textbf{Correct output:}};
        \node[text=black, inner sep=8pt, minimum width=40pt, minimum height=25pt] at (1*\spacing,-1) {False};

         \node[anchor=east] at (7.1,-1) {\textbf{Model's output:}};
        \node[ text=red, inner sep=8pt, minimum width=40pt, minimum height=25pt] at (6*\spacing,-1) {\textbf{True}};
        %\node[text=black, inner sep=8pt, minimum width=50pt, minimum height=25pt] at (3*\spacing,-1) {\textbf{Model's output:} {\color{red} True}};

\hidden{
        % Add a line after "Output: True"
        \node[anchor=east] at (-0.8,-1) {\textbf{Model's output:}};
        \node[ text=red, inner sep=8pt, minimum width=40pt, minimum height=25pt] at (1*\spacing,-1) {True};
        \node[anchor=east] at (-0.8,-2) {\textbf{Correct output:}};
        \node[ text=black, inner sep=8pt, minimum width=40pt, minimum height=25pt] at (1*\spacing,-2) {False};
        }
    \end{tikzpicture}
    \caption{Attention maps for an ICL example for the task "every" of type $(x_1,f(x_1),x_2,f(x_2),...,x_n)$, where the query $x_n$ is a big value. Dark blue indicates weights approaching 0, while yellow 
indicates weights approaching 1.}\label{attn-map}
\end{figure*}

%% ─────────────────────────────────────────────
\section{The Softmax Problem}\label{sec:softmax}
%% ─────────────────────────────────────────────

%\noindent\textbf{Purpose.} 
In this section, we show both mathematically and empirically 
why the Softmax scoring function is a primary cause of ICL generalization failures under distribution shift and the attention collapse observed in the previous section.

\vspace{0.5em}
\noindent\textbf{Softmax saturation.} To recall the basics of attention, 
let $\text{e}^\ell=(\text{e}_1^\ell,\ldots,\text{e}_n^\ell)$ be the 
input embeddings processed by the multi-head attention mechanism at 
layer $\ell$, where $\text{e}_i^\ell$ denotes the embedding of the 
$i$-th token. Each attention head $h$ in layer $\ell$ is parameterized 
by Query, Key, and Value weight matrices $\text{Q}^{h,\ell}$, 
$\text{K}^{h,\ell}$, and $\text{V}^{h,\ell}$, with $\text{d}_k$ the 
embedding size divided by the number of heads. The output of each head 
is:
%\begin{equation}\label{attention}
%\small
\[
\text{C}_i^{h,\ell} = \sum^n_{j=1} \softmax\left(
\frac{(\text{Q}^{h,\ell}\text{e}_i^\ell)^{\top}
(\text{K}^{h,\ell}\text{e}_j^\ell)}{\sqrt{\text{d}_k}}
\right) \text{V}^{h,\ell}\text{e}_j^\ell
%\end{equation}
\]

Once the gap between values in the Softmax argument exceeds a threshold, the distribution concentrates on its maximum. Let $z_j$ denote the logits and assume $z_{j_0}=\max_j z_j$. Writing $\Delta_j = z_{j_0}-z_j \ge 0$, we have
\[
\softmax(z_{j_0}) = \frac{1}{1+\sum_{k\neq j_0} e^{-\Delta_k}}
\]
\[
\softmax(z_j) = \frac{e^{-\Delta_j}}{1+\sum_{k\neq j_0} e^{-\Delta_k}}
\]
If $\Delta_j \ge \delta$ for all $j\neq j_0$, then
\[
\softmax(z_{j_0}) \ge \frac{1}{1+(n-1)e^{-\delta}}
\]
\[
\softmax(z_j) \le e^{-\Delta_j} \le e^{-\delta}.
\]
Thus, for $\delta \approx 4$ (so $e^{-4}\approx 0.018$), the maximal weight is close to $1$ while all others are negligible, and the output collapses to:
\begin{equation}\label{final}
\text{C}_{i}^{h,\ell} = \text{V}^{h,\ell}\text{e}_{j_0}^\ell
\end{equation}
making the attention head's representation of the token dependent only on a context of size one. 
Importantly, the large values $x_j$ yielding Equation \ref{final} come 
from inputs the model has seldom seen in training; the $\text{Q}$ and 
$\text{K}$ matrices were never trained to handle them.

%{\color {magenta} we need a discussion here for what happens with in distribution vs. out of distribution, because softmax performs very well in distribution.}

\vspace{0.5em}
\noindent\textbf{Empirical confirmation.} Following standard practice 
for ICL in small transformers, we use a linear embedding 
$\mathrm{emb}: x \mapsto x \cdot W$, for $(x,W) \in \mathbb{R} \times 
\mathbb{R}^d$. This embedding preserves magnitude ordering: if $|x| < 
|y|$ then $\|\mathrm{emb}(x)\| < \|\mathrm{emb}(y)\|$. As a direct 
consequence, a large input value will always produce a large embedding norm, making Softmax saturation not merely possible but inevitable 
when deviant elements are present. Figure~\ref{attn-map} confirms this 
prediction directly: when the sequence contains the value $70$, the 
attention layer assigns virtually all weight to that token, ignoring 
the elements that actually determine the truth value of \ref{every}. 
The model consequently predicts the sequence as all positive, based 
solely on the large value. With significant differences in input values, 
Softmax increasingly resembles Hardmax, assigning weight close to 1 to 
the largest element and near 0 to all others. This concentration effect 
is not limited to linear embeddings: significant norm differences also 
arise with linguistic tokens, as seen in the \textit{OpenWebText} corpus 
(Figure~\ref{token}, Appendix~\ref{appendix:token}).

\vspace{0.5em}
\noindent\textbf{Effect on linear functions.} Softmax saturation 
adversely affects the linear function task as well. For example, a 12L8AH model predicting $f(x) = x$ with input sequence $[100, -1.09, 
0.78, 0.26, 0.42]$ produces predictions $[-1.21, -0.28, 2.15, 0.96, 
0.65]$, a complete failure to approximate the 
function.%\footnote{Though eventually the model begins to recover and approximate better.} 

When a value $x_i$ in the sequence input to the attention mechanism is larger than the other elements of the sequence and other elements in its training, Softmax will assign $x_i$ probability $1$ and all other elements in the sequence probability $0$. This makes sense in some tasks; a large value in the attention mechanism intuitively signals a strong statistical correlation in context sensitive aspects of meaning \cite{asher:2011}; Softmax amplifies this value.  However, in tasks like ours this is problematic. 
An input with a large norm representing a large number does not necessarily have a disproportionately greater effect.  % is not necessarily disproportionately greater. 
%in our task might have a big semantic value in comparison to other inputs, its importance might not be all that much greater for the task than the other smaller tokens. 
For our tasks, the model must look at many tokens in the context; with deviant sequences, Softmax prevents the models from doing this.    

\hidden{
Training on wider distributions such as $D_{\cal I} = {\mathcal 
N}(0,10)$ or $D_{\cal I} = {\mathcal N}(0,100)$ partially mitigates 
the problem on deviant sequences, but significantly degrades performance 
on ${\mathcal N}(0,\sigma)$ for small $\sigma$, as the $\text{Q}$ and 
$\text{K}$ matrices learn smaller weights to compensate. Moreover, all 
models still fail once out-of-distribution elements become sufficiently large. {\color{blue} TODO}
}
\vspace{0.5em}
\noindent\textbf{Takeaway.} Softmax saturation is a fundamental 
architectural limitation: it causes attention to collapse onto a single 
token whenever large-magnitude inputs are present, regardless of whether 
that token is relevant to the task. This is particularly harmful for 
ICL, which requires integrating information across many context tokens 
simultaneously.

\section{Exploring Alternatives to Softmax}\label{sec:alternatives-main}

%\noindent\textbf{Purpose.} 
Having identified Softmax saturation as the 
source of generalization failures, a natural question is whether 
existing alternative scoring functions already address this problem.  In this section we answer that this is not the case.

\vspace{0.5em}

We systematically evaluated a broad range of alternatives: 
temperature-scaled Softmax with $\tau \in \{5,10,20,50,100\}$ or a trainable temperature parameter; 
sparse attention mechanisms (Sparsemax~\cite{sparsemax} and 
Entmax~\cite{entmax}); hybrid schemes partitioning attention heads 
across tanh, uniform averaging, ReLU, and $x^2$ scoring functions 
(details in Appendix~\ref{sec:alternatives}); and linear or modified 
Softmax approaches, namely linear attention, CosFormer~\cite{qin:etal:2022} and 
SA-Softmax~\cite{zheng:etal:2025}. None of these yielded consistent improvements over standard Softmax on our tasks.

\vspace{0.5em}

\noindent\textbf{Takeaway.} Despite the breadth of alternatives tested, 
none succeeded in mitigating the saturation problem identified in the 
previous section. This suggests that a more targeted solution is needed, 
which we develop in the next section.

 \begin{figure}[ht!]  
%\centering
%\includegraphics[width=0.49\textwidth]{figures/les_mse.jpeg}
\includegraphics[width=0.5\textwidth]{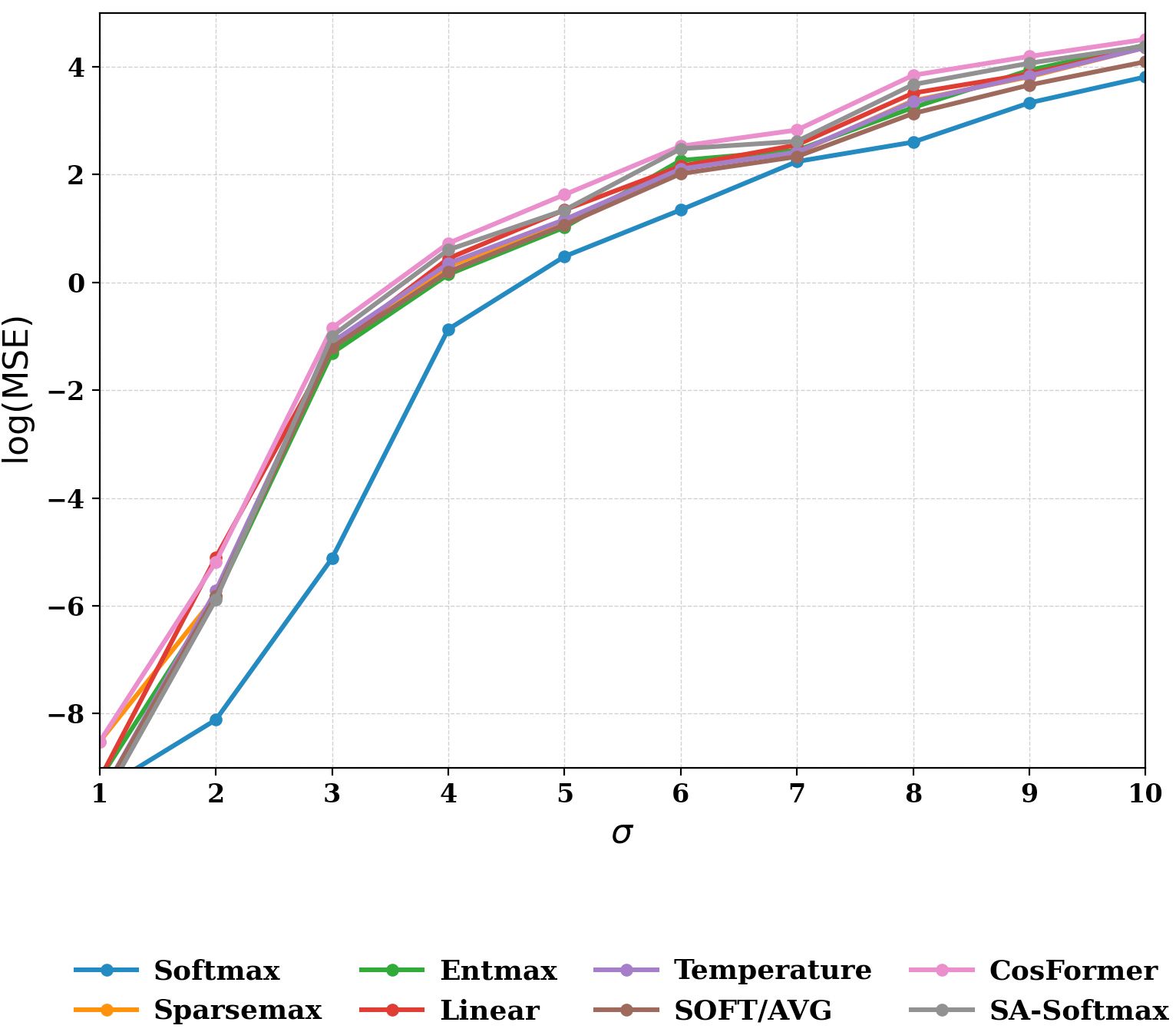}

\caption{Evolution of $\log(\text{MSE})$ across models tested on $x \in D^t_{\cal I} = {\mathcal N}(0,1)$ and weights $a,b \in D^t_{\cal F} = {\mathcal N}(0,\sigma)$, as a function of the distribution shift parameter $\sigma$. All models use a 12-layer, 8-head full transformer. Exact values are reported in Table~\ref{table:LF} in Appendix \ref{sec:alternatives}.
\label{alternatives}}
\end{figure}

\section{Solution: Signed Scaled Averaging (SSA)}

For ICL tasks that require integrating information across many tokens, Softmax saturation is clearly harmful. Conversely, tasks requiring focused attention on specific tokens can benefit from it. Rather than relying on a one-size-fits-all scoring function, we propose a mechanism that can adaptively control the degree of selectivity.

\paragraph{Selective Scaled Attention (SSA).}
To improve ICL performance, we replace the exponential scoring function with a parameterized alternative applied elementwise:
\[
x \;\mapsto\; (1 + b |x|)^{\operatorname{sgn}(x)\, n},
\]
where $b > 0$ and $n \ge 1$ are trainable parameters.

The absolute value ensures a positive base, while the signed exponent induces asymmetric behavior: positive inputs grow polynomially, whereas negative inputs decay toward zero.

For a vector $z=(z_1,\dots,z_p)\in\mathbb{R}^p$, SSA is defined as
\begin{equation}
\mathrm{SSA}(z)_i \;=\; 
\frac{(1 + b\,|z_i|)^{\operatorname{sgn}(z_i)\,n}}
     {\sum_{k=1}^{p} (1 + b\,|z_k|)^{\operatorname{sgn}(z_k)\,n}}
\end{equation}

\paragraph{Interpolation between regimes.}
SSA defines a continuous family interpolating between polynomial and exponential behaviors. For $x \ge 0$, setting $b = \tfrac{1}{m}$ and $n = m$ yields
\[
(1 + b x)^n = \left(1 + \tfrac{x}{m}\right)^m \xrightarrow[m \to \infty]{} e^x,
\]
recovering exponential growth. At the other extreme, for $x \ge 0$ and when $b = n = 1$, we obtain
\[
(1 + b x)^n = 1 + x,
\]
which is linear. 

\paragraph{Contextualization: SSA vs Softmax}
SSA differs from Softmax under two key transformations of the logits. The first is
 global magnification: the logits are multiplied by a common factor, which can
occur through the embedding or under out-of-distribution inputs in our ICL
tasks. The second is inner disproportion: one token has an increasingly larger
score than the others like in the experiments in Figure~\ref{attn-map}.
%In this case, Softmax rapidly suppresses tokens of
%secondary importance; the theorem below measures this saturation speed. 

In both cases ---magnification and inner disproportion--- SSA tends to keep a more balanced
view of context, whereas Softmax rapidly tends towards hardmax, even in the presence of temperature.

In this paragraph, $b$ and $n \ge 1$ are fixed, so is the SSA map; similarly, we fix a temperature $\tau>0$ for Softmax. Since non-positive logits rapidly vanish, we restrict to positive logits.

We first address the magnification question, where we see that Softmax saturates very fast while SSA keeps a more balanced view of very large logits.

\begin{theorem}[Logit magnification]\label{thm:scale-separation}
Let $r\in\mathbb{R}^k$ be a positive vector and let $i_\ast=\argmax\{r_i:i=1,\ldots,k\}$ be a maximizing token, assumed to be unique.  
Given a scaling factor $\lambda>0$, consider the two attention distributions
\[
    \alpha_i^{\mathrm{SM},\tau}(\lambda)
    =
    \frac{e^{\lambda r_i/\tau}}{\sum_j e^{\lambda r_j/\tau}},
\]
and
\[
    \alpha_i^{\mathrm{SSA}}(\lambda)
    =
    \frac{(1+b\lambda r_i)^n}{\sum_j(1+b\lambda r_j)^n},
\]
for $i=1,\ldots,k$. 
Then, as $\lambda\to\infty$,
\[
    \alpha_i^{\mathrm{SM},\tau}(\lambda)\to
    \begin{cases}
    1, & i=i_\ast,\\
    0, & i\neq i_\ast,
    \end{cases}
\]
while
\[
    \alpha_i^{\mathrm{SSA}}(\lambda)\to
    \frac{r_i^n}{\sum_j r_j^n}.
\]
In other words, the residual ``contextual mass'' satisfies
\begin{align*}
    &1-\alpha_{i_\ast}^{\mathrm{SM},\tau}(\lambda)\to0,\\
&  1-\alpha_{i_\ast}^{\mathrm{SSA}}(\lambda)\to
    1-\frac{r_{i_\ast}^n}{\sum_j r_j^n} ,\quad \mbox{ as }\lambda\to\infty.
\end{align*}
\end{theorem}

\paragraph{Proof.}
For Softmax, divide all terms by $e^{\lambda r_{i_\ast}/\tau}$.  If
$j\neq i_\ast$, then
\[
    \frac{e^{\lambda r_j/\tau}}{e^{\lambda r_{i_\ast}/\tau}}
    =
    e^{-\lambda(r_{i_\ast}-r_j)/\tau}
    \to0,
\]
because $r_{i_\ast}>r_j$.  Hence all nonmaximizing tokens vanish and
$\alpha_{i_\ast}^{\mathrm{SM},\tau}(\lambda)\to1$, while
$\alpha_j^{\mathrm{SM},\tau}(\lambda)\to0$ for $j\neq i_\ast$.

For SSA, since every $r_i$ is positive,
\[
    (1+b\lambda r_i)^n
    =
    (b\lambda)^n
    \left(r_i+\frac{1}{b\lambda}\right)^n .
\]
The common factor $(b\lambda)^n$ cancels in the normalization.  The remaining
terms converge to $r_i^n$, giving the stated limit. $\Box$

We now study inner disproportion, understood here as the collapse of contextual mass when
the imbalance between the relative importance or score of tokens increases.

\begin{theorem}[Contextual mass collapse]\label{th:saturation-speed}
Let $s=(s_1,\ldots,s_k)$ in $\mathbb{R}^k$ be a positive vector  and
let $i_\ast$ be the index of the maximizing token. Assume that $\rho:=s_{i_\ast}\ge1$ and
$s_j\le1$ for all $j\neq i_\ast$.
Then temperature-scaled Softmax satisfies
\[
    1-\alpha_{i_\ast}^{\mathrm{SM},\tau}(s)
    \le (k-1)e^{-(\rho-1)/\tau},
\]
whereas SSA satisfies
\[
    1-\alpha_{i_\ast}^{\mathrm{SSA}}(s)
    =
    \frac{\sum_{j\neq i_\ast}(1+bs_j)^n}
    {(1+b\rho)^n+\sum_{j\neq i_\ast}(1+bs_j)^n}
\]
and so
\[
    1-\alpha_{i_\ast}^{\mathrm{SSA}}(s)
    =
    O(\rho^{-n}).
\]

\end{theorem}

\paragraph{Proof.}
For Softmax, since $s_j\le1$ for $j\neq i_\ast$,
\[
1-\alpha_{i_\ast}^{\mathrm{SM},\tau}(s)
\le
\sum_{j\neq i_\ast} e^{-(\rho-s_j)/\tau}.
\]
Since $s_j\le1$ for all $j\neq i_\ast$,
\[
\sum_{j\neq i_\ast} e^{-(\rho-s_j)/\tau}
\le
(k-1)e^{-(\rho-1)/\tau}.
\]
For SSA, the displayed formula follows directly from the attention weights. Since
$s_j\le1$ for $j\neq i_\ast$, the numerator is bounded by
$(k-1)(1+b)^n$, giving the stated polynomial bound. $\Box$

In conclusion, as the size of the winning token comes to dominate, temperature-scaled Softmax tends to hardmax exponentially fast, while SSA moves only polynomially, allowing for a more subtle contextualization.

Additional theoretical analysis of SSA is provided in Appendix~\ref{appendix:ssa-theory}.

\vspace{0.3cm}
\noindent\textbf{Takeaway.}
%\jer{
Theorems 1 and 2 show, from two different perspectives, that SSA keeps a stronger awareness of contextual variety.  Under global magnification, Softmax loses the contextual mass outside the winning token, whereas SSA preserves a distribution shaped by the relative scores. Under inner disproportion, Softmax moves to hardmax exponentially fast, whereas SSA collapses only polynomially. This matters for ICL, where the prediction often depends on several contextual examples rather than on a single dominant token. SSA therefore does not only soften attention: it gives each head a trainable way to delay collapse and keep secondary tokens active when the context still matters, while introducing only two scalar parameters.

\section{Evaluating SSA}\label{sec:ssa-eval}

\begin{figure*}[ht!]
%\includegraphics[width=11cm]{figures/evolution of errorsbis.png} 
%\center 
%\includegraphics[width=6.9cm]{figures/ANDSOFTMAX.png}
%\includegraphics[width=6.9cm]{figures/ANDSSA.png}
%\includegraphics[width=6.5cm]{figures/andsooftmax.png}
%\includegraphics[width=7cm]{figures/ANDSSAA.png}
%\includegraphics[width=5.8cm]{figures/ANDSOFTMAX_2.png}
%\includegraphics[width=6.5cm]{figures/ANDSSA_2.png}

%\includegraphics[width=0.49\textwidth]{figures/ANDSOFTMAX_2.png}
%\includegraphics[width=0.49\textwidth, height=5.5cm]{figures/ANDSSA_2.png}

%\includegraphics[width=1\textwidth]{figures/errrate.png}
\makebox[\textwidth][c]{%
        \includegraphics[width=1.08\textwidth]{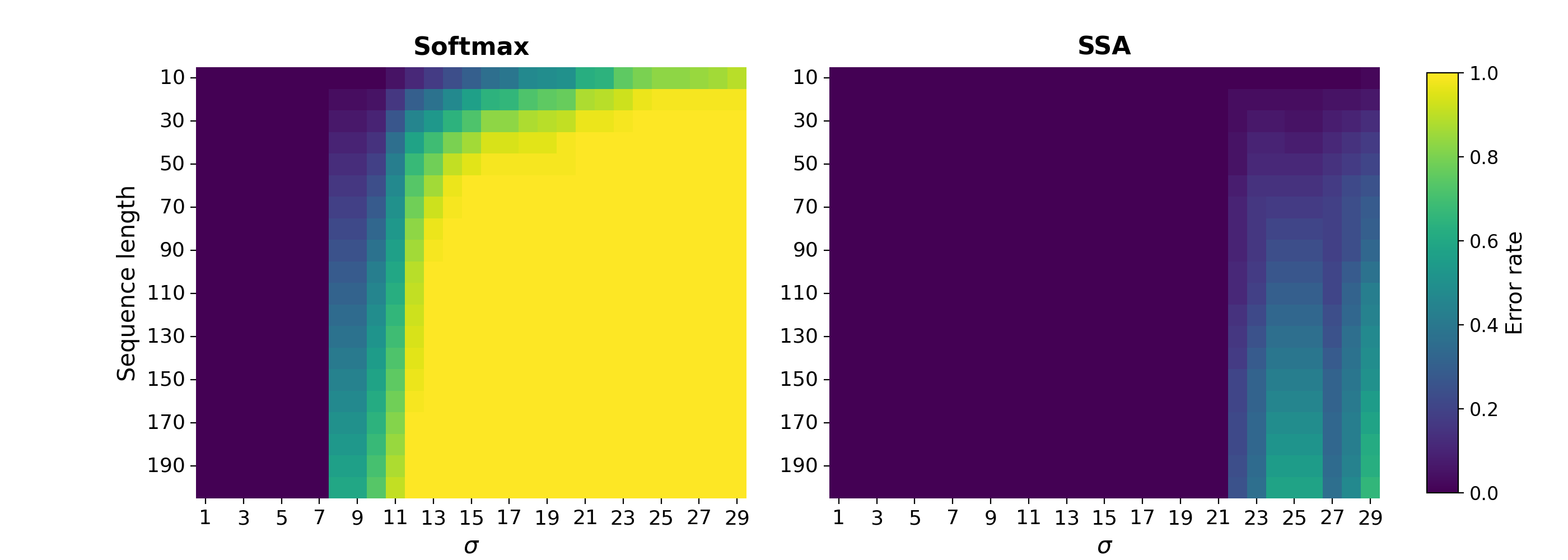}
    }

\caption{Heatmaps showing the evolution of errors for the 12L8AH model with Softmax (Left) and SSA (Right) on the  \emph{every} task. Model was trained on data in $D_{\cal I}={\mathcal N}(0,1)$ for lengths from 11 to 40 and tested in $D^{test}_{\cal I}={\mathcal N}(0,\sigma)$ for $\sigma \in \{1,...,30\}$ and lengths from 10 to 200 for each task.   Yellow represents a much higher error rate than purple.}
\label{hmap1}
\end{figure*}

 \begin{figure*}[ht!]  
\centering
\hidden{
\includegraphics[width=0.48\textwidth, height=5.58cm]{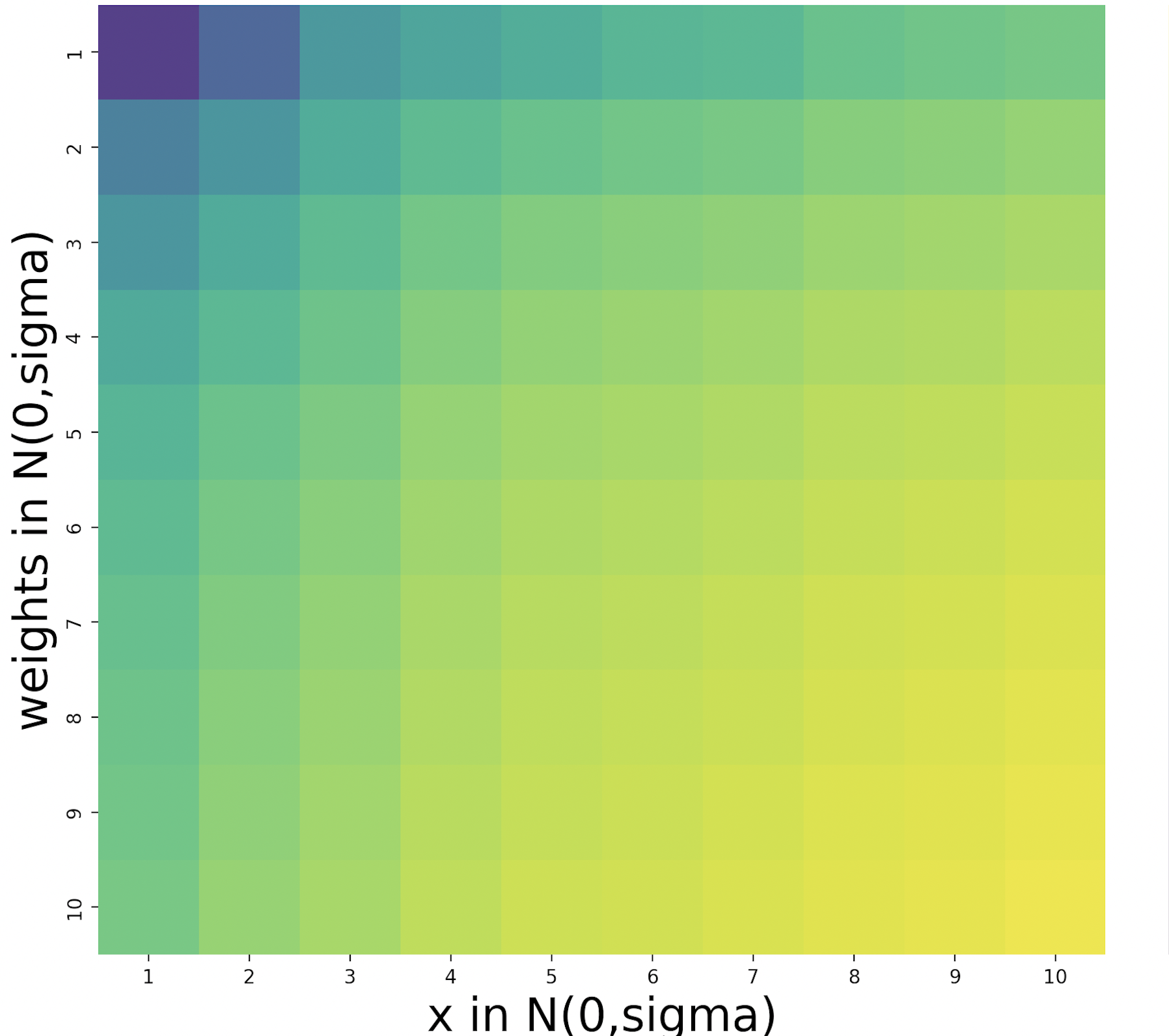}
\includegraphics[width=0.48\textwidth]{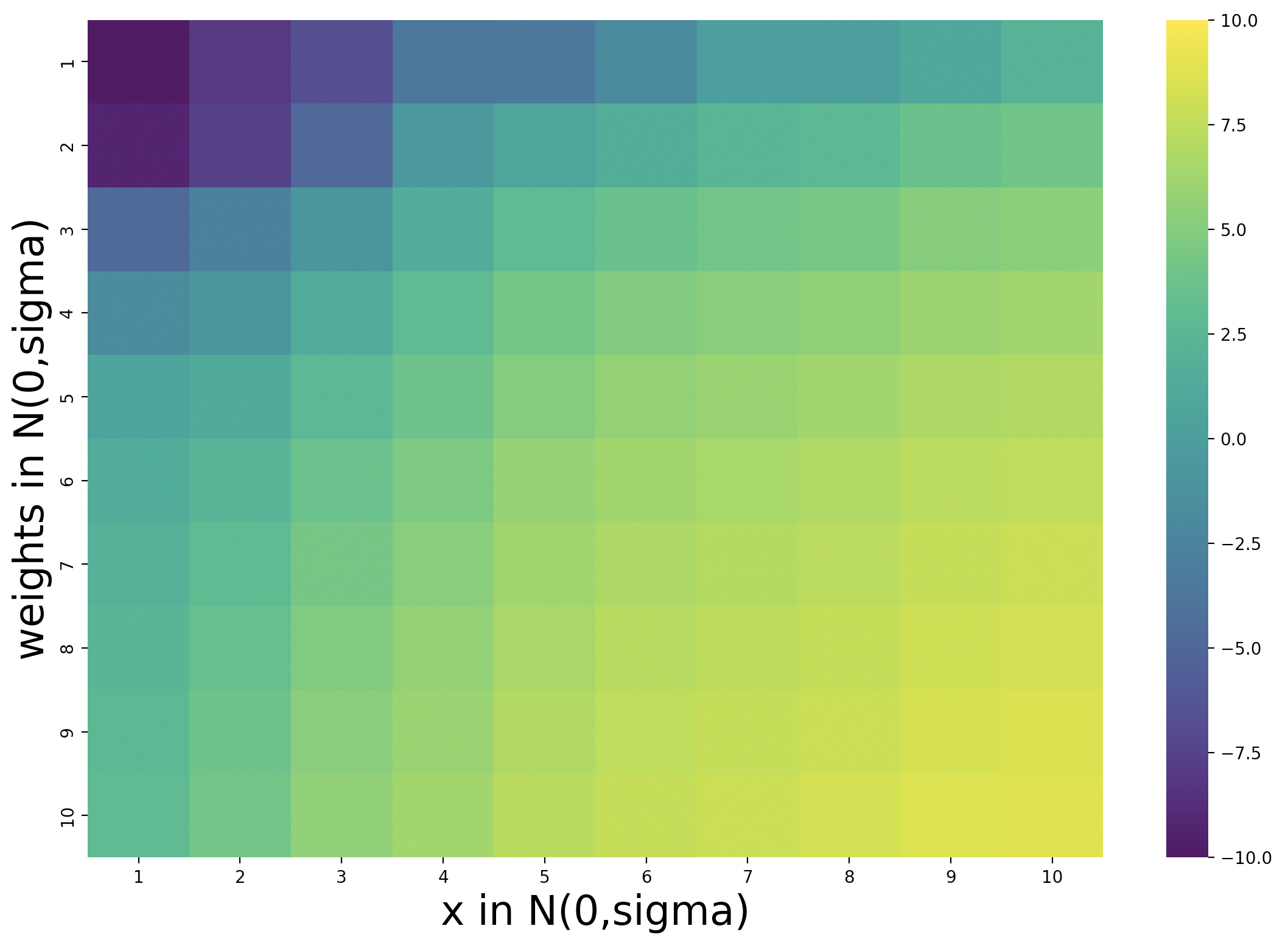}
}

\makebox[\textwidth][c]{%
        \includegraphics[width=1.08\textwidth]{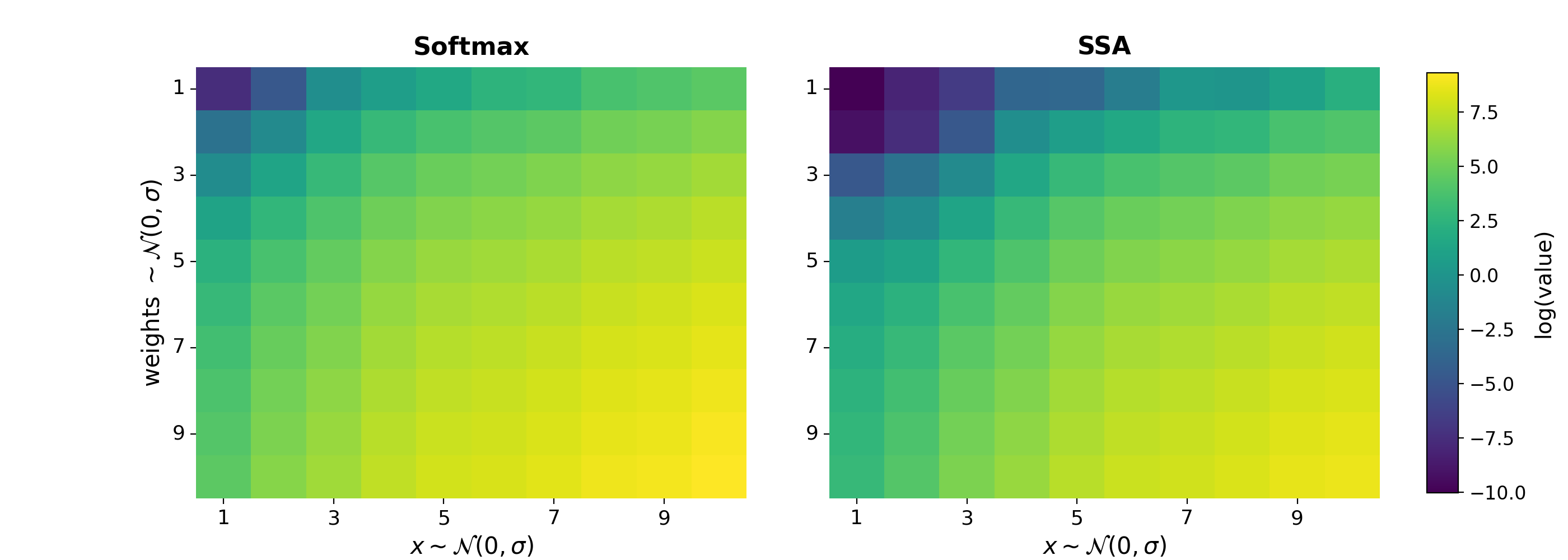}
    }
\caption{ (Left) Comparison plot showing the evolution of log MSE for SSA and Softmax-based models (12L8AH) with $D_{\cal F} , D_{\cal I} , D^{test}_{\cal I} \sim {\mathcal N}(0,1)$ and varying $D_{\cal F}^{test} \sim {\mathcal N}(0, \sigma)$. The heatmap shows the evolution of  logarithm of MSE for the Softmax (Left) and SSA (Right) model when varying both $D^{test}_{\cal I}$ and $D^{test}_{\cal F}$. \label{solution}}
\end{figure*}

%\noindent\textbf{Purpose.} 
We now show how SSA's theoretical 
advantages translate into empirical gains, first on the controlled ICL 
tasks that motivated its design, then on broader NLP benchmarks to test 
whether the benefits generalize beyond synthetic settings.

%% ──────────────────────────────────────────
\subsection{Effects of SSA on ICL Tasks}
%% ──────────────────────────────────────────

Substituting SSA for Softmax substantially improves performance on both 
ICL tasks. All results use our 12L8AH full transformer model. For the 
quantification task, SSA yields considerable generalization improvement 
both in handling longer test sequences and in managing deviant inputs 
with extreme values. Figure~\ref{hmap1} compares error rates for the 
\emph{every} task: the SSA model (right) maintains lower error across 
broader ranges of sequence length and input distribution compared to the 
Softmax model (left). SSA also improves performance on the \emph{some} 
task (Appendix~\ref{sec:appendixG}). For the linear function task, SSA 
substantially improves performance when tested on out-of-distribution 
function parameters (Figure~\ref{solution}), outperforming all 
alternatives tested (Table~\ref{table:LF}).

\vspace{0.5em}
\noindent\textbf{Takeaway.} SSA directly addresses the failure modes 
identified in Section~\ref{sec:softmax}: by avoiding saturation, it 
allows attention to remain distributed across context tokens even in 
the presence of large-magnitude inputs. We next ask whether this 
advantage persists in real NLP settings.

%% ──────────────────────────────────────────
\subsection{SSA on Decoder-Only NLP Benchmarks}
%% ──────────────────────────────────────────

We trained a Nemotron-style decoder-only model (114M parameters, 12 
layers, 24 attention heads, hidden dimension 768) from scratch on 10B 
tokens from the FineWeb corpus for 22k steps , using a 
custom tokenizer with a vocabulary of 50,256 tokens. We compared a 
standard Softmax baseline against an SSA variant under identical 
training conditions, and evaluated both models on a range of benchmarks using the LM Eval harness \cite{lmeval}.

As shown in Table~\ref{tab:softmax_vs_ssa}, SSA consistently outperforms Softmax across all evaluated tasks. Table~\ref{tab:perplexity} further shows that SSA achieves lower perplexity on both in-distribution FineWeb data and out-of-distribution Wikipedia text, with training loss curves reported in Figure~\ref{fig:loss_curves} (Appendix~\ref{lossssa}).

\begin{table}[ht]
\centering
\scriptsize
\begin{tabular}{llrr}
  \toprule
  \textbf{Benchmark} & \textbf{Metric} & \textbf{Softmax} & \textbf{SSA} \\
  \midrule
%  arc\_challenge   & acc          & $0.2048 \pm 0.0118$ & $\mathbf{0.2355 \pm 0.0124}$ \\
  arc\_challenge   & acc\_norm    & $0.2398 \pm 0.0125$ & $\mathbf{0.2713 \pm 0.0130}$ \\
%  arc\_easy        & acc          & $0.2959 \pm 0.0094$ & $\mathbf{0.5547 \pm 0.0102}$ \\
  arc\_easy        & acc\_norm    & $0.2934 \pm 0.0093$ & $\mathbf{0.5387 \pm 0.0102}$ \\
  boolq            & acc          & $0.3783 \pm 0.0085$ & $\mathbf{0.5618 \pm 0.0087}$ \\
  cb               & acc          & $0.1429 \pm 0.0472$ & $\mathbf{0.4643 \pm 0.0672}$ \\
  cb               & f1           & $0.1310$             & $\mathbf{0.2663}$             \\
  copa             & acc          & $0.5900 \pm 0.0494$ & $\mathbf{0.6400 \pm 0.0482}$ \\
 % hellaswag        & acc          & $0.2595 \pm 0.0044$ & $\mathbf{0.2947 \pm 0.0045}$ \\
  hellaswag        & acc\_norm    & $0.2550 \pm 0.0043$ & $\mathbf{0.3283 \pm 0.0047}$ \\
  multirc          & acc          & $0.4280 \pm 0.0071$ & $\mathbf{0.4350 \pm 0.0071}$ \\
 % openbookqa       & acc          & $0.1280 \pm 0.0150$ & $\mathbf{0.2060 \pm 0.0181}$ \\
  openbookqa       & acc\_norm    & $0.2860 \pm 0.0202$ & $\mathbf{0.3040 \pm 0.0206}$ \\
  record           & f1           & $0.1983 \pm 0.0040$ & $\mathbf{0.2482 \pm 0.0043}$ \\
  record           & em           & $0.1932 \pm 0.0039$ & $\mathbf{0.2427 \pm 0.0043}$ \\
  wic              & acc          & $0.5000 \pm 0.0198$ & $\mathbf{0.5078 \pm 0.0198}$ \\
  winogrande       & acc          & $0.4972 \pm 0.0141$ & $\mathbf{0.5178 \pm 0.0140}$ \\
  \bottomrule
\end{tabular}
\caption{Zero-shot benchmark comparison between Softmax and SSA on a 
Nemotron-style decoder model (114M) trained for 22K steps on FineWeb. 
Bold indicates the better result.}
\label{tab:softmax_vs_ssa}
\end{table}

\begin{table}[ht]
\centering
\small
\begin{tabular}{lrr}
  \toprule
  \textbf{Evaluation} & \textbf{Softmax} & \textbf{SSA} \\
  \midrule
  FineWeb (in-dist.) $\downarrow$    & 21.86 & \textbf{19.73} \\
  Wikipedia (out-dist.) $\downarrow$ & 24.58 & \textbf{22.07} \\
  \bottomrule
\end{tabular}
\caption{Perplexity comparison between Softmax and SSA 
(114M Nemotron-style decoder, 22K steps). Bold indicates 
the better result.}
\label{tab:perplexity}
\end{table}

%\vspace{0.5em}
\noindent\textbf{Takeaway.} SSA's benefits are not confined to synthetic tasks: it achieves lower perplexity on both in-distribution and out-of-distribution text, lower training loss, and consistent gains across a diverse set of NLP benchmarks, all after only 22k steps of training on FineWeb.

%% ──────────────────────────────────────────
\subsection{SSA on Encoder-Only Models}
%% ──────────────────────────────────────────

To test SSA beyond decoder-only architectures, we trained variants of 
BabyBERTa \citep{huebner:etal:2021} on the AO-CHILDES corpus, comparing 
the standard Softmax baseline with SSA using fixed exponents $n=1.5$ 
and $n=2$. BabyBERTa is a compact RoBERTa-style encoder trained on 
child-directed speech, well-suited for probing grammatical dependency 
learning in limited-data settings. We evaluated the models on linguistic 
probes from \cite{huebner:etal:2021} using the masked language modeling 
(MLM) metric, which measures token-level accuracy against distractors.

%\begin{table}[h!]
%\small{
%\centering
%\begin{tabular}{|l|ccc|}

\begin{table}[ht]
\centering
\small
\begin{tabular}{llrr}
\toprule
\textbf{linguistic probe} & \textbf{Softmax} & \textbf{SSA 1.5} & \textbf{SSA 2} \\
\midrule
%\textbf{agreement\_Det\_N-across\_1\_adj}       & 75.35 & 73.5 & 73.3 \\
agr\_subj\_verb-across\_PP & 56 & 58.95 & {\bf 65.95} \\
agr\_subj\_verb-across\_RC & 55.5 & 57.7 & {\bf 61.55} \\
agr\_subj\_verb-in\_Q+aux & 76.5 & {\bf 79.0} & 70.45 \\
anaphor\_agr-pron\_gender & 48.1 & 51.45 & {\bf 53.7} \\
arg\_str-dropped\_arg & 79.65 & 74.65 & {\bf 85.55} \\
arg\_str-swapped\_args & 83.3 & {\bf 92.0} & 88.0 \\
arg\_str-transitive & 53.44 & 53.85 & {\bf 57.2} \\
binding-principle\_a & 78.25 & {\bf 87.9} & 80.2 \\
case-subjective\_pron & 85.55 & 89.7 & {\bf 91.75} \\
%\textbf{ellipsis-n\_bar} & {\bf 60.75} & 53.3 & 53.45 \\
%\textbf{filler-gap-wh\_question\_object} & {\bf 91.7} & 89.4 & 87.15 \\
filler-gap-wh\_Q\_subject & 79.2 & {\bf 83.3} & 68.25 \\
irregular-verb & 70.05 & {\bf 78.3} & 69.2 \\
%\textbf{local\_attractor-in\_question\_with\_aux} & 85.45 & 81.85 & 85.0 \\
%\textbf{quantifiers-existential\_there} & {\bf 91.3} & 85.25 & 76.25 \\
quantifiers-superlative & 71.2 & {\bf 83.95} & 65.25 \\ %\hline
%\textbf{OVERALL} & 73.01 & {\bf 74.94} & 72.48
\bottomrule
\end{tabular}

\caption{
Performance of BabyBERTa models trained from scratch on AO-CHILDES using Softmax and SSA (1.5, 2), evaluated on linguistic probes from \cite{huebner:etal:2021} with the MLM metric.
%BabyBERTa Model performance trained from scratch on AO-CHILDES with Softmax and two settings of SSA, evaluated with the MLM metric on various linguistic probes  from \cite{huebner:etal:2021}. 
Agr: agreement; arg\_str: argument structure; Subj: subject; Pron: pronoun; PP: prepositional phrase; RC: relative clause; Det: determiner; N: noun; arg: argument.
} \label{table:MLM}
\end{table}

As shown in Table~\ref{table:MLM}, SSA improves BabyBERTa's performance 
across a range of grammatical phenomena. SSA-2 achieves the strongest 
gains on syntax-sensitive tasks such as subject--verb agreement across 
relative clauses and argument structure alternations, while SSA-1.5 
yields improvements on morphological and lexical tests such as irregular 
verbs and pronoun gender.

\vspace{0.5em}
\noindent\textbf{Takeaway.} Consistent with the decoder-only results, 
SSA enhances grammatical sensitivity in encoder-only models as well. 
Taken together, the results across controlled tasks, decoder models, 
and encoder models support SSA as an effective drop-in replacement for 
Softmax across transformer architectures.

\section{Conclusion}
Transformer models %  have severe problems in generalizing to out of distribution data, something we have also empirically confirmed.
struggle to generalize effectively to out-of-distribution data.
%, a limitation from prior research that we have also empirically confirmed.
We identified the Softmax scoring function in the attention mechanism as a factor contributing to this challenge and introduced SSA, a novel scoring method  %SSA substantially improves the performance of small transformers on both mathematical and NLP tasks by better capturing linguistic structure, all without increasing model complexity. Its parametrizable and adaptable nature makes SSA a versatile alternative to the standard Softmax, demonstrating clear advantages over a one-size-fits-all approach.
%contributor to this issue and 
%We have isolated one cause of these problems in the Softmax scoring function of the attention mechanism.  We 
%proposed a new type of scoring function, scaled signed averaging (SSA),
that significantly improves the performance 
%of small transformer models
on both mathematical and NLP tasks. 

SSA enhances a model’s ability to capture linguistic structure and allocate attention more effectively, addressing Softmax’s tendency to saturate and focus narrowly on a few tokens. By parametrically controlling attention, SSA improves generalization without increasing model complexity or reducing accuracy. While it does not solve all generalization challenges, SSA directly mitigates those caused by Softmax’s inflexible scoring, offering a simple yet effective drop-in alternative for transformers' attention mechanism.

\section*{Limitations}

Due to hardware and data constraints, we were unable to scale models beyond 114M parameters. We trained a Nemotron-style decoder-only model (114M parameters, 12 layers, 24 attention heads, hidden dimension 768)  from scratch on 10B tokens from the FineWeb corpus for 22k steps, using a custom tokenizer with a vocabulary of 50,256 tokens. 
This required approximately 2 days on 12 NVIDIA A100 80GB GPUs per model. 
To enable a rigorous comparison, we trained both a Softmax and an SSA version under identical conditions, resulting in a total of over 4 days of compute.

While SSA provides consistent improvements in generalization, it does not fully resolve all limitations of the attention mechanism. In particular, our results indicate that SSA still struggles in scenarios involving strong distribution shifts, where both the input $x_i$ and the test-time function distribution $D^{\text{test}}_{\mathcal{F}}$ differ significantly from the training distribution $D_{\mathcal{F}}$.

%In addition, SSA does not address the fact that, as we have noted above, the simple mathematical structure of the attention mechanism conflates the value of tokens  with their importance for the particular task. 

More fundamentally, this points to a broader open question regarding the attention mechanism itself: its underlying mathematical structure tends to conflate the magnitude of token representations with their relevance to the task. As a result, tokens with large values can disproportionately influence attention, even when they are not the most informative for the prediction. Addressing this misalignment remains an important direction for future work.

%This and our observations about boundary values provide further empirical support for the induction head hypothesis.

%training, with the projection values bounded by the model's boundary values.   %It finds sequences $\vec{y_i}$ within some distance $d$ of the sequence $\vec{x}$ in the prompt for the target function and the induction heads use the values $y_i_n}$ to determine a value for $f(x_n)$, given $x_n$.  

 %We hope our work will inspire further research into what transformer-based models are actually doing on ICL tasks.

%\section*{Author Contributions}
%If you'd like to, you may include  a section for author contributions as is done
%in many journals. This is optional and at the discretion of the authors.

\section*{Acknowledgments}

This work was supported by SARER and Summ-RE projects (ANR-20-CE23-0017), and the AI Cluster ANITI (ANR-19-PI3A-0004). Computational resources were provided by CALMIP under Grant 2016-P23060.

%Use unnumbered first level headings for the acknowledgments. All
%acknowledgments, including those to funding agencies, go at the end of the paper.
\hidden{
\section*{Ethics Statement}
Our studies show potentially important consequences for the use of both fine-tuned and prompted models.  To apply Softmax transformer models with good guarantees to tasks, one needs to know either their pre-training (in the case of prompting) or at least the distributions used for fine-tuning.  Outside of those ranges, the models can quickly cease to make reliable predictions. 
}

\bibliography{custom}

\appendix

\section{Training details}
\label{sec:appendixA}

\subsection{ICL Tasks}
Our ICL tasks involve training from scratch on sequences containing in-context examples (input-output pairs) $(x_1, f(x_1), ..., x_i)$ ending with a query input $x_i$ that is used to generate the corresponding output.

\paragraph{Model Architecture.} 
We use a decoder-only Transformer architecture from the GPT-2 family~\citep{radford2019language} with 12 layers, 8 attention heads, and a 256-dimensional embedding space. Dropout is set to 0 as we sample fresh prompts at each training step. The model takes as input a sequence of vectors in its embedding space and predicts the next vector in the sequence.

\paragraph{Input/Output Encoding.} 

Both prompt inputs and outputs are then mapped into the model's latent embedding space of dimension 256 through a learnable linear transformation $W_{\text{enc}} \in \mathbb{R}^{256}$. The model processes this sequence and outputs vectors in the same embedding space. These output vectors are mapped back to scalar predictions via a separate learnable linear transformation $W_{\text{dec}} \in \mathbb{R}^{256}$ (implemented as a dot product). %The model prediction at position $2i-1$ corresponds to the prediction of $g(x_i)$, and by design, this prediction depends only on $(x_j, g(x_j))$ for $j < i$ and $x_i$. Predictions at positions corresponding to $g(x_i)$ are ignored during training. This architecture allows computing predictions for all prompt prefixes in a single forward pass.

\paragraph{Training Procedure.} 
Models are trained for 500k steps using the Adam optimizer~\citep{diederik2014adam} with a learning rate of $10^{-4}$ and batch size of 64. At each training step, we sample a fresh batch of random prompts by: (1) sampling a random function $g$ from the function class according to $\mathcal{D}_\mathcal{F}$, (2) sampling inputs $x_1, \dots, x_{k+1}$ independently from $\mathcal{D}_\mathcal{I}$, and (3) evaluating $g$ on these inputs to produce the prompt. For each prompt, the loss is computed as $\frac{1}{k}\sum_{i=1}^{k} (\hat{y}_i - g(x_i))^2$ where $\hat{y}_i$ is the model's prediction.

\paragraph{Curriculum Learning.} 
We conduct training both with and without curriculum learning. When employing curriculum learning, we train on a set $S$ of training sequences of varying lengths, ranging from 1 to $k=40$. Specifically, we start with prompt length 3 (number of input-output pairs). Every 2,000 training steps, we increase the length by 2, until reaching the full prompt length.

\hidden{
\subsection{Linear function training details:}
{\color{blue} To reformulate}
To train the model to ICL ${\cal L}$, we looked for a $\theta^{*}$ that optimizes the following auto-regressive objective: $$ \theta^{*} = \argmin _\theta \mathbb{E}_{x_i \in D_{\cal I} , f \in D_{\cal F}}$$$$\small{\left[ \sum_{i=0}^k l
\left(f\left(x_{i+1}\right), {\cal L}_\theta\left((x_1,f(x_1),...,f(x_i), x_{i+1})\right)\right)\right]} $$
where ${\cal L}_\theta$ is a ``learner'', $l$ is squared error and $f : x \rightarrow ax + b$ is a linear function with $a,b$ sampled from some training distribution for functions. $D_{\cal F}$ and samples $x_i$ are sampled from  a training distribution for points $D_{\cal I}$.  We will note that $f \in D_{\cal F}, x \in D_{\cal I}$. We choose at random a function $f \in D_{\cal F}$ and then a sequence of points $x_i \in D_{\cal I}$ as random prompts, from a distribution $D_{\cal I}$ at each training step.  We update the model through a gradient update. We use a batch size of 64 and train for 500k steps. The models saw over 1.3 billion training examples for each distribution we studied. For $D_{\cal F}$ and $D_{\cal I}$ we used several distributions: the normal distribution ${\mathcal N}(0,1)$,  ``rectangle" or uniform distributions over given intervals and bimodal distributions.  
%\large{\bf Appendix A: Training details} 
}
\paragraph{Additional training information:} We use the Adam optimizer \cite{diederik2014adam} , and a learning rate of $10^{-4}$ for all models.\\

\paragraph{Computational resources:} We used Nvidia A-100 GPUs to train the different versions of transformer models from scratch, with an average training time of 4 hours and used Nvidia Volta (V100 - 7,8 Tflops DP) GPUs for the fine-tuning of LLaMA 3.1 8B involved in these experiments.

\paragraph{ Fine-tuning} LLaMA 3.1 8B was fine tuned on 26000 randomly generated sequences $S$ progressing from 1 to $40$ for 2 epochs using LoRA \citep{lora}.  The input values were drawn from ${\mathcal N}(0,1)$. At test time, we were only able to run a few sequences for each possible we examined sequence lengths from 10 to 200 and number distributions from ${\cal N}(0,\sigma)$ for $1 \leq \sigma \leq 30$. This meant that we could only run a few test batches, since we need to look at 600 total pairs for each task.  We averaged the prediction errors on the each $(S^{\text{test}}, D^{\text{test}}_{\cal I})$ possibility.  

 For our attempted fine-tuning of LLaMA 3.1 8B on the function task, we set the input sequence to be of the form $(x_1,f(x_1),x_2,f(x_2),....,x_n)$ requiring that the output be of the form $f(x_n)$, a single numerical value.  The model returned a list of values 
 $(w_1,w_2,...)$.  It failed to capture the basic input and output pattern.

\subsection{Training on FineWeb}
\label{finewebtraining}
\paragraph{Training Details.}
We trained a Nemotron3-style Transformer decoder model from scratch with customized dimensions, totaling 114M parameters. The model consists of 12 Transformer layers with 24 attention heads and grouped-query attention (GQA) with 8 query groups, a hidden dimension of 768, a feed-forward hidden dimension of 3{,}072, and a maximum context length of 1{,}024 tokens. Input and output embeddings are shared, and the vocabulary size is 50{,}256 tokens. Training was performed using the Adam optimizer with $\beta_1 = 0.9$, $\beta_2 = 0.95$, $\varepsilon = 10^{-5}$, and a weight decay of $0.1$. We used a peak learning rate of $3 \times 10^{-4}$ with a cosine decay schedule over the full training horizon, computed from the total token budget of approximately 10B tokens. The global batch size was 128 sequences with a micro-batch size of 1 and a sequence length of 1{,}024 tokens, yielding approximately $131$k tokens per step. Mixed-precision training was conducted in BFloat16 using a distributed optimizer. For SSA, the parameter $b$ was fixed to 0.8, while $n$ was learned as a per-layer parameter (shared across attention heads).

\hidden{
\section{SSA settings}
For each task, we trained two classes of models differing only in the scoring function. The first uses the standard Softmax-based attention mechanism. The second replaces Softmax with \emph{SSA}. In the SSA setting, parameters are learned for each attention head, such that every head in every Transformer layer has its own set of SSA parameters. Aside from the choice of scoring function and its parameterization, all architectural components, optimization settings, and training procedures were kept identical across models.
}
Aside from the choice of scoring function and its parameterization, all architectural components, optimization settings, and training procedures were kept identical across models. As a result, we observe no meaningful difference in training time between SSA- and Softmax-based models.

\section{SSA settings}
For each task, we trained two classes of models differing only in the scoring function. The first uses the standard Softmax-based attention mechanism, while the second replaces Softmax with \emph{SSA}. In the SSA setting, a small number of additional parameters are learned independently for each attention head, such that every head in every Transformer layer has its own set of SSA parameters. Concretely, SSA introduces only two scalar parameters per attention head. For a model with 12 layers and 8 attention heads per layer, this corresponds to a total of $2 \times 12 \times 8 = 192$ additional parameters, which is negligible compared to the overall model size. All initialization and optimization details required for exact reproducibility are provided in the released code. Aside from the choice of scoring function and its parameterization, all architectural components, optimization settings, and training procedures were kept identical across models.

%, i.e. the model outputs a list of values. 
% We did not know how to evaluate such a list. %

%\newpage
%The error rate we chose is: $r_\epsilon = \frac{\epsilon_\sigma}{|\epsilon_* - \epsilon_0|}$ where $\epsilon_*$ is the best $\epsilon_\sigma$ error for a model M with $\fh(x)$ calculated with Least Squares, and $\epsilon_0$ is the worst $\epsilon_\sigma$ error for a model $M$ such that $\fh_M(x) = 0$, $\forall x$.  %In all error calculations, we exclude the first two predictions of each batch from the squared error calculation.  %We need at least two points to be able to find a linear function and the first two predictions by the model are hence almost always wrong. %A drawback of this method is that if a batch gives abnormal predictions, its error will be diluted by all the other batches and we won't notice it. %So we also calculated an average error $\delta$ over batches.

\section{Using mixtures of scoring functions} \label{sec:alternatives}

\begin{table*}[!ht]
\centering
\small{
\begin{tabular}{|l|l|l|l|l|l|l|l|l|l|l|}
 \hline
 %\rowcolor{orange!30}
 models \ $\backslash$ \ $\sigma$ & 1 & 2 & 3 & 4 & 5 & 6 & 7 & 8 & 9 & 10 \\ 
 \hline\hline
%  SSA   & $4\times10^{-5}$& $3\times10^{-4}$& $10^{-3}$& $0.02$& $0.02$ & $0.15$& $1.24$& $1.04$& $2.74$& $8.50$ \\

 Softmax   &$8\times 10^{-5}$& $3\times10^{-4}$& $6\times10^{-3}$& $0.42$& $1.62$& $3.84$& $9.42$& $13.51$& $27.99$& $45.35$ \\
 Sparsemax & $2\times10^{-4}$& $3\times10^{-3}$& 0.28& 1.35& 3.07& 8.73& 11.06& 29.15& 45.39& 78.16 \\
 Entmax & $1\times10^{-4}$& $2.9\times10^{-3}$& 0.27& 1.16& 2.78& 9.62& 11.60& 25.66& 51.30& 81.74 \\

Linear &	$10^{-4}$	& $6.1\times10^{-3}$ &	0.3 &	1.56 &	3.84	& 8.64 &	12.82 &	33.54	& 47.86 & 79.06 \\
Trainable $\tau$ & $7\times 10^{-5}$ & $3.3\times 10^{-3}$ & 0.33 &  1.42 & 3.21 & 8.21 &  11.22 & 28.60& 46.22 & 78.02
\\
 $\tau=5$  & $3\times10^{-4}$ & $1\times10^{-2}$ & 0.81 & 3.17 & 7.37 & 16.33 & 21.43 & 51.28 & 74.55 & 103.71 \\
$\tau=10$ & $1\times10^{-4}$ & $3\times10^{-3}$ & 0.29 & 1.25 & 2.98 & 7.61  & 10.43 & 23.43 & 41.96 & 68.81  \\
$\tau=20$ & $5.2\times10^{-5}$ & $2.9\times10^{-3}$ & 0.30 & 1.22 & 2.97 & 8.05 & 11.25 & 28.48 & 47.13 & 78.53  \\
$\tau=50$ & $1\times10^{-4}$ & $3.1\times10^{-3}$ & 0.29 & 1.29 & 3.18 & 8.14 & 11.68 & 24.15 & 42.81 & 73.51  \\

$\tau=100$ & $1\times10^{-4}$ & $2.5\times10^{-3}$ & 0.26 & 1.11 & 2.82 & 8.53 & 12.07 & 29.02 & 51.43 & 80.49  \\

 SOFT/AVG & $7\times10^{-5}$& $3\times10^{-3}$& $0.30$& $1.22$& $2.91$& $7.52$& $10.32$& $22.97$& $38.97$& $60.03$ \\
 4 Scoring fts &  $5\times10^{-5}$ &  $3\times10^{-3}$ & $0.34$& $1.33$& $3.18$& $8.28$& $10.99$& $26.31$&$ 42.42$& $70.33$ \\
 CosFormer &  $2\times10^{-4}$ & $5.6 \times10^{-3}$& $0.43$ & $2.07$ & $5.10$ & $12.54$ & $16.91$ & $46.68$ & $66.22$ & $91.02$ \\
 SA-Softmax & $6\times 10^{-5}$& $2.8\times 10^{-3}$& $0.37$& $1.83$& $3.83$& $11.94$& $13.72$& $39.29$& $58.43$& $80.31$ \\
  Llama 3.3 70b  &  $2\times10^{-3}$& $5.50$& $3.16$& $13.71$& $18.21$& $23.91$& $28.99$& $33.51$& $40.25$& $48.02$ \\ [1ex] 
 \hline
\end{tabular}
}
\caption{Comparison showing the evolution of squared errors for models tested on $x \in D^t_{\cal I} = {\mathcal N}(0,1)$ and weights $a,b \in D^t_{\cal F}={\mathcal N}(0,\sigma)$. Temperature-scaled Softmax is evaluated with either a fixed temperature $\tau \in \{5,10,20,50,100\}$ or a trainable temperature parameter learned during training. All models use a 12-layer Transformer with 8 attention heads.}
\label{table:LF}
\end{table*}

A natural test is to take temperature-scaled Softmax with  parameter~$\tau$ to rescale the exponential behavior. We tried several scaling factors $ \tau \in \{5, 10, 20, 50, 100\}$, but none produced better results than the standard Softmax. 
In addition, we experimented with alternative normalization functions that yield sparse attention distributions, including Sparsemax~\cite{sparsemax} and 
%$1.5$-
Entmax~\cite{entmax}. These methods replace the Softmax normalization while preserving the overall attention framework. However, neither Sparsemax nor $\alpha$-Entmax led to performance improvements on our tasks.

We next partitioned the attention heads such that half utilized Softmax-based scoring, while the remaining half employed uniform averaging over all tokens. This design, we thought, would preserve contextual breadth, reduce the risk of focusing to specific tokens, and also increase the model’s expressiveness through multiple scoring functions.
We experimented with four distinct known scoring functions (tanh, average, ReLU, and $x^2$), assigning two heads to each (For detailed results see Table \ref{table:LF}). This approach improved over than Softmax on ${\cal N}(0,1)$ but was less good elsewhere. 

We additionally tested \textsc{cosFormer}~\cite{qin:etal:2022}, which replaces the exponential weighting in Softmax with a cosine-modulated kernel combined with linear normalization. But, cosFormer did not improve performance on our ICL tasks and often underperformed the Softmax baseline (Table~\ref{table:LF}).

Finally, we experimented with the recently proposed Self-Adjusting Softmax (SA-Softmax)~\cite{zheng:etal:2025}. However, this mechanism did not yield better generalization or accuracy than the standard Softmax attention (Table~\ref{table:LF}).

\section{Additional Supporting Analysis for the SSA Function}
\label{appendix:ssa-theory}

Let's consider:
\begin{equation}
    f(x) = (1 + b|x|)^{\,\operatorname{sgn}(x)\,n},
    \qquad b>0,\; n\ge1,
\end{equation}

Logits are denoted $s_1,\ldots,s_k \in \mathbb{R}$.

\begin{equation}
    \alpha_i^{\text{SSA}}
    =
    \frac{f(s_i)}{\sum_{j=1}^k f(s_j)}.
\end{equation}

In this section, we provide additional analytical details for the function $f$, including its $C^1$-regularity.

% ------------------------------------------------------------------
\subsection{Growth Behavior: SSA vs.\ Exponential}

Softmax uses the exponential function, whose super-polynomial growth causes severe
sensitivity to logit differences:
\[
    e^x \to \infty \quad \text{exponentially in } x.
\]

SSA behaves fundamentally differently.  
Since $\operatorname{sgn}(x)$ determines the exponent, SSA admits the piecewise form
\[
f(x) =
\begin{cases}
    (1+bx)^{n}, & x\ge0, \\[4pt]
    (1+b|x|)^{-n}, & x<0.
\end{cases}
\]

Hence the asymptotics are
\[
x\to +\infty:\quad f(x)\sim (bx)^n, \]
\[ x\to -\infty:\quad f(x)\sim (b|x|)^{-n}.
\]

Thus $f$ grows only polynomially, in stark contrast to the exponential.  
It therefore reacts more gently to large positive logits.   

\begin{figure}[!ht]

\includegraphics[width=\columnwidth]{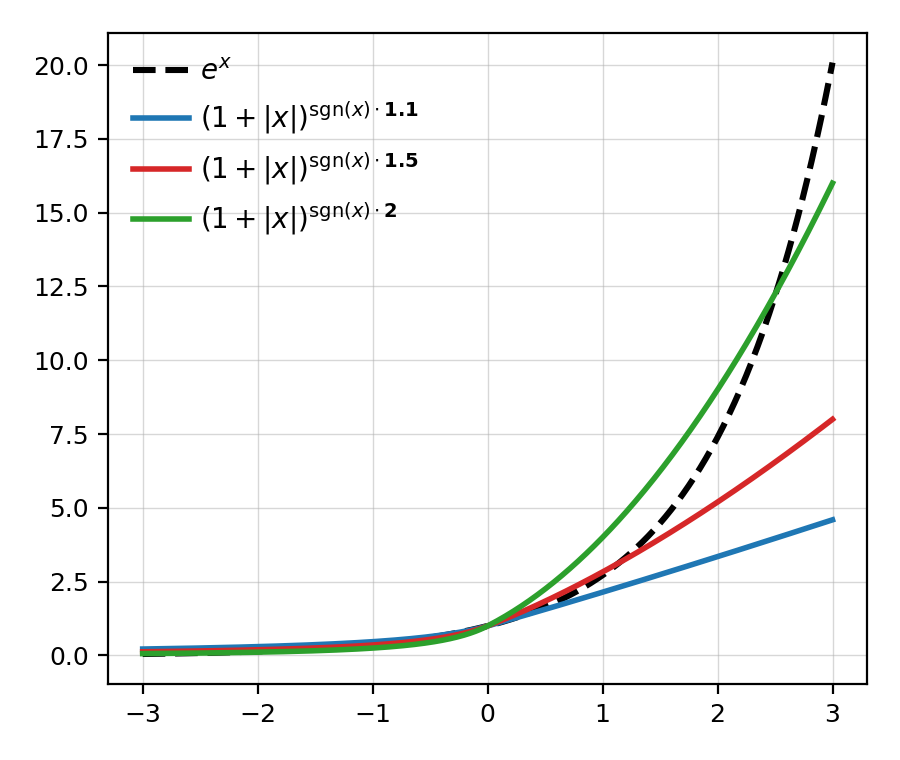}
\caption{
Illustration of the base function $(1 + b|x|)^{\mathrm{sgn}(x)n}$ used in SSA, 
plotted for $b=1$ and $n \in \{1.1, 1.5, 2\}$. 
The curves demonstrate that SSA exhibits behavior similar to the exponential function, 
but with a tunable growth rate toward \(+\infty\), which increases with larger values of the exponent \(n\).
}
\label{exSSA}
\end{figure}
% ------------------------------------------------------------------
\subsection{Gradient Structure and Logarithmic Slope}

\paragraph{Piecewise derivative.}
Differentiation yields
\[
f'(x)=
\begin{cases}
    nb(1+bx)^{\,n-1}, & x>0,\\[4pt]
    nb(1-bx)^{-n-1}, & x<0, \\
    nb,  & x=0,
\end{cases}
\]
Therefore $f$ is $C^1$ and
\[
|f'(x)| =
\begin{cases}
    \Theta(|x|^{n-1}),   & x\to+\infty,\\[4pt]
    \Theta(|x|^{-n-1}), & x\to -\infty.
\end{cases}
\]
Softmax satisfies $f'(x)=e^x$, which diverges exponentially for large $x$.

\paragraph{Logarithmic slope.}
Define $g(x)=\frac{f'(x)}{f(x)}$.  
From the expressions above, 
\[
    g(x)=\frac{nb}{1+b|x|},
\]

The graph of g(x) for ssa shows that extremely large inputs are ``cropped''; an increase in the size of large input will only provide a small increase in score, whereas for small values SSA is quite sensitive. This does not occur with Softmax with $g(x) = 1$: large inputs give large outputs.

\section{Heat map for the "some" task with SSA for model trained from scratch}
\label{sec:appendixG}
\begin{figure}[!ht]
\center 

\includegraphics[width=\columnwidth]{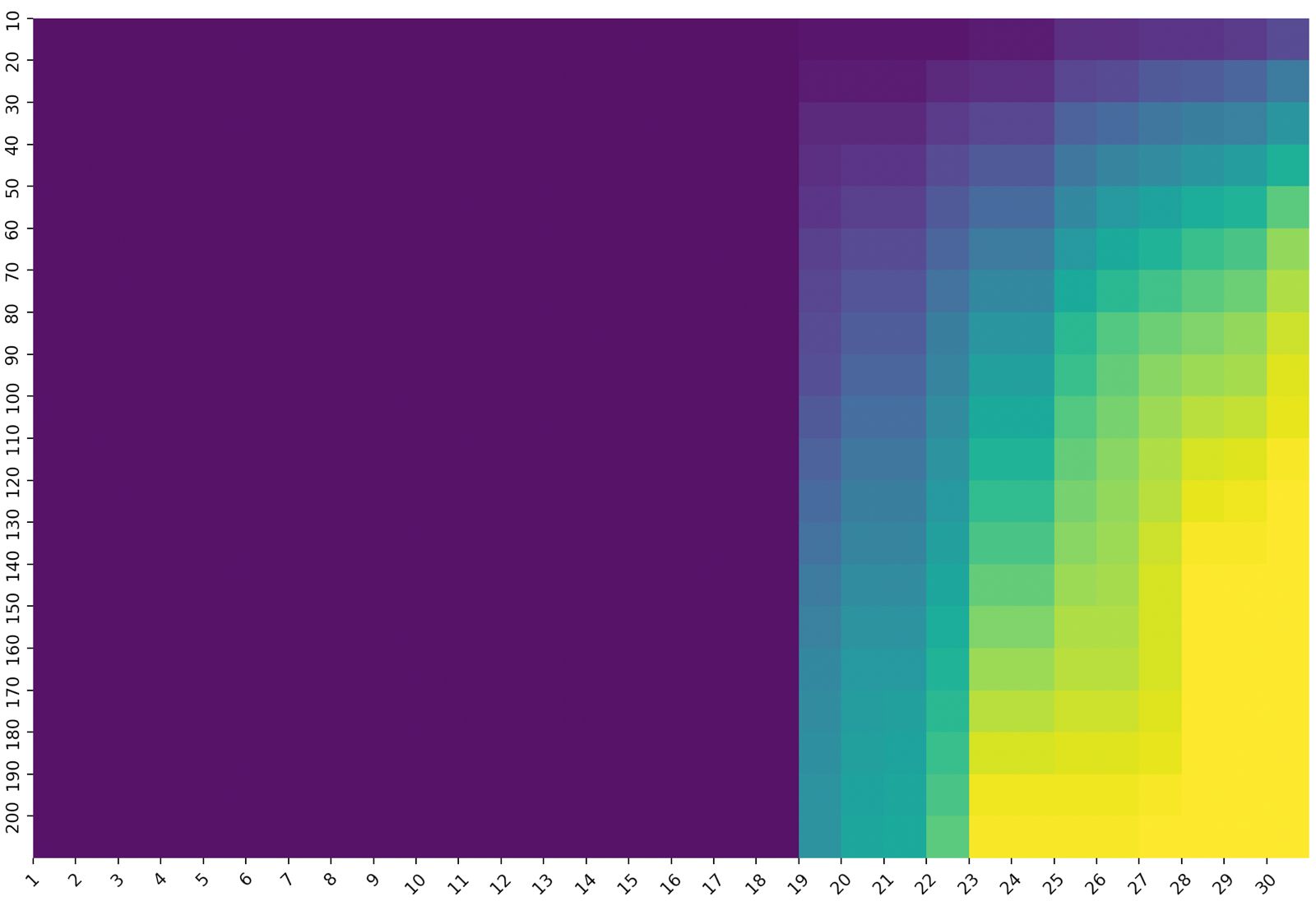}
\includegraphics[width=\columnwidth]{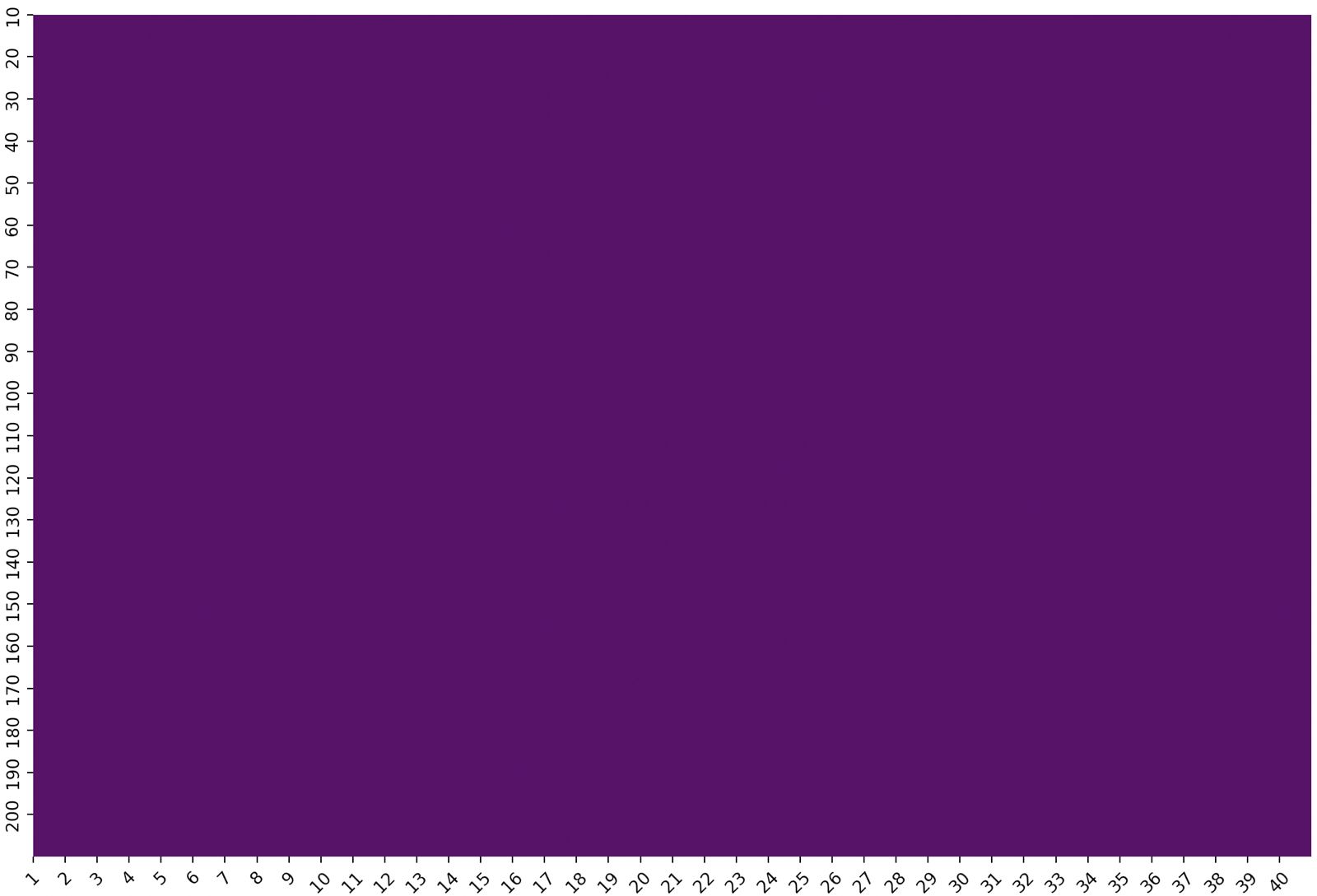}

\caption{Heatmap showing the evolution of errors for the task \emph{some}  trained on data in $D_{\cal I}={\mathcal N}(0,1)$ for lengths from 11 to 40 and tested in $D^{\text{test}}_{\cal I}={\mathcal N}(0,\sigma)$ for $\sigma \in \{1,...,30\}$ and lengths from 10 to 200. The first figure is for the Softmax-based model and the second with SSA. Yellow represents a much higher error rate than purple.} 
\label{hmap2}
\end{figure}
%The figure of this appendix is \ref{p2>p1}
%\large{\bf Appendix C: Failure to generalize to longer prompt sequences}

\newpage

\section{Individual Sign Classification Under Distribution Shift}
\label{app:sign-classification}

To confirm that generalization failures on the quantification task are not caused 
by the model losing the ability to represent or classify individual values, we 
tested whether the model can correctly determine the sign of a single 
out-of-distribution input. For each $\sigma \in \{1, \ldots, 29\}$, we presented 
the model with a one-point prompt where the input was drawn from 
$\mathcal{N}(0, \sigma)$ and the ground truth label was the sign of that input.

\begin{figure}[ht!]
    \centering
    \includegraphics[width=\columnwidth]{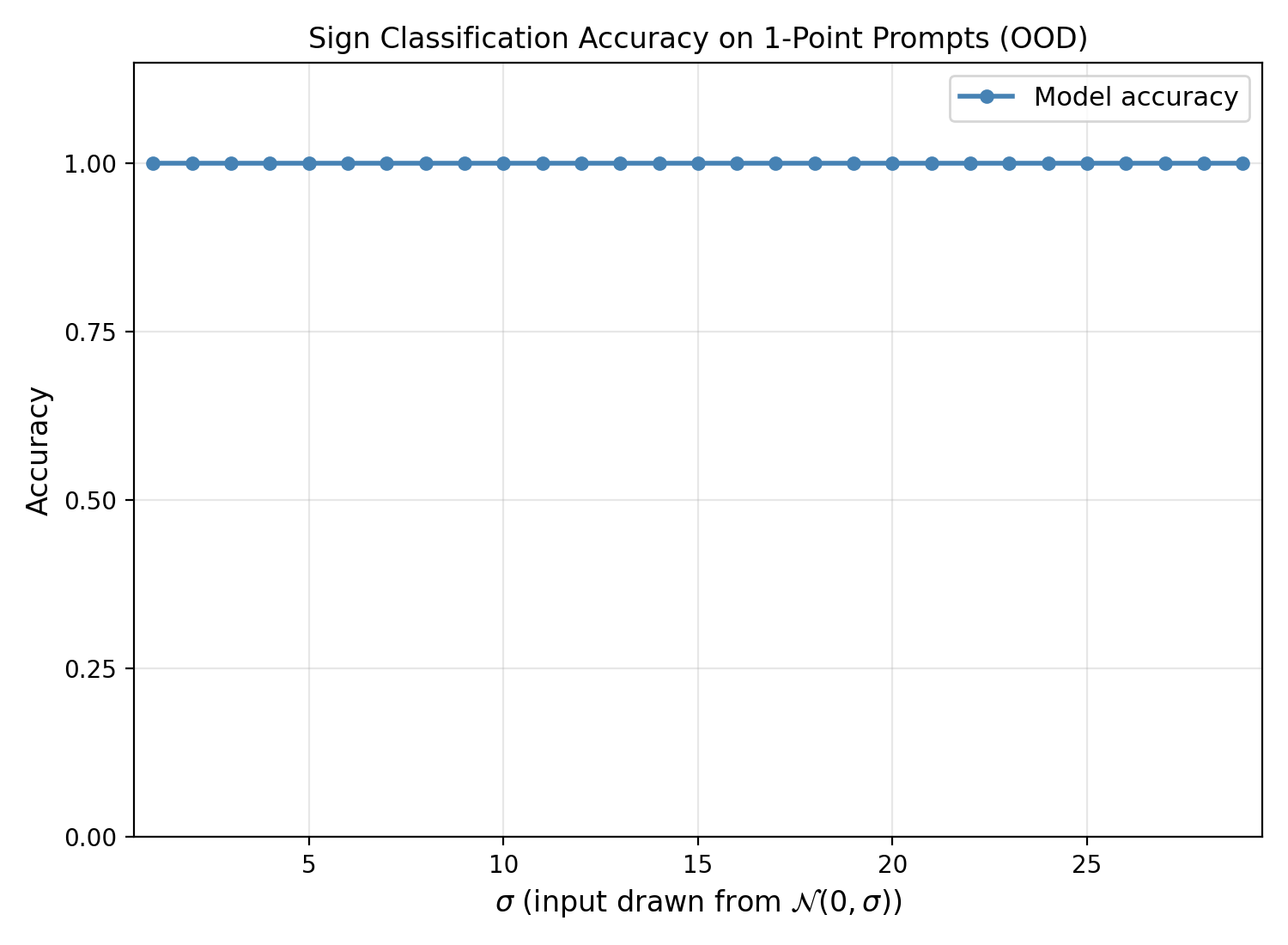}
    \caption{Sign classification accuracy of the 12L8AH Softmax model on 
    one-point prompts with inputs drawn from $\mathcal{N}(0, \sigma)$, for 
    $\sigma \in \{1, \ldots, 29\}$.}
    \label{fig:sign-classification}
\end{figure}

As shown in Figure~\ref{fig:sign-classification}, the model achieves perfect 
accuracy (1.0) across all values of $\sigma$, with zero errors out of 64 
sequences in every condition, regardless of how far the input lies outside the 
training distribution $\mathcal{N}(0,1)$.

\newpage
\section{Training Loss Evolution: SSA vs Softmax}
\label{lossssa}

\begin{figure}[ht]
    \centering
    \includegraphics[width=\columnwidth]{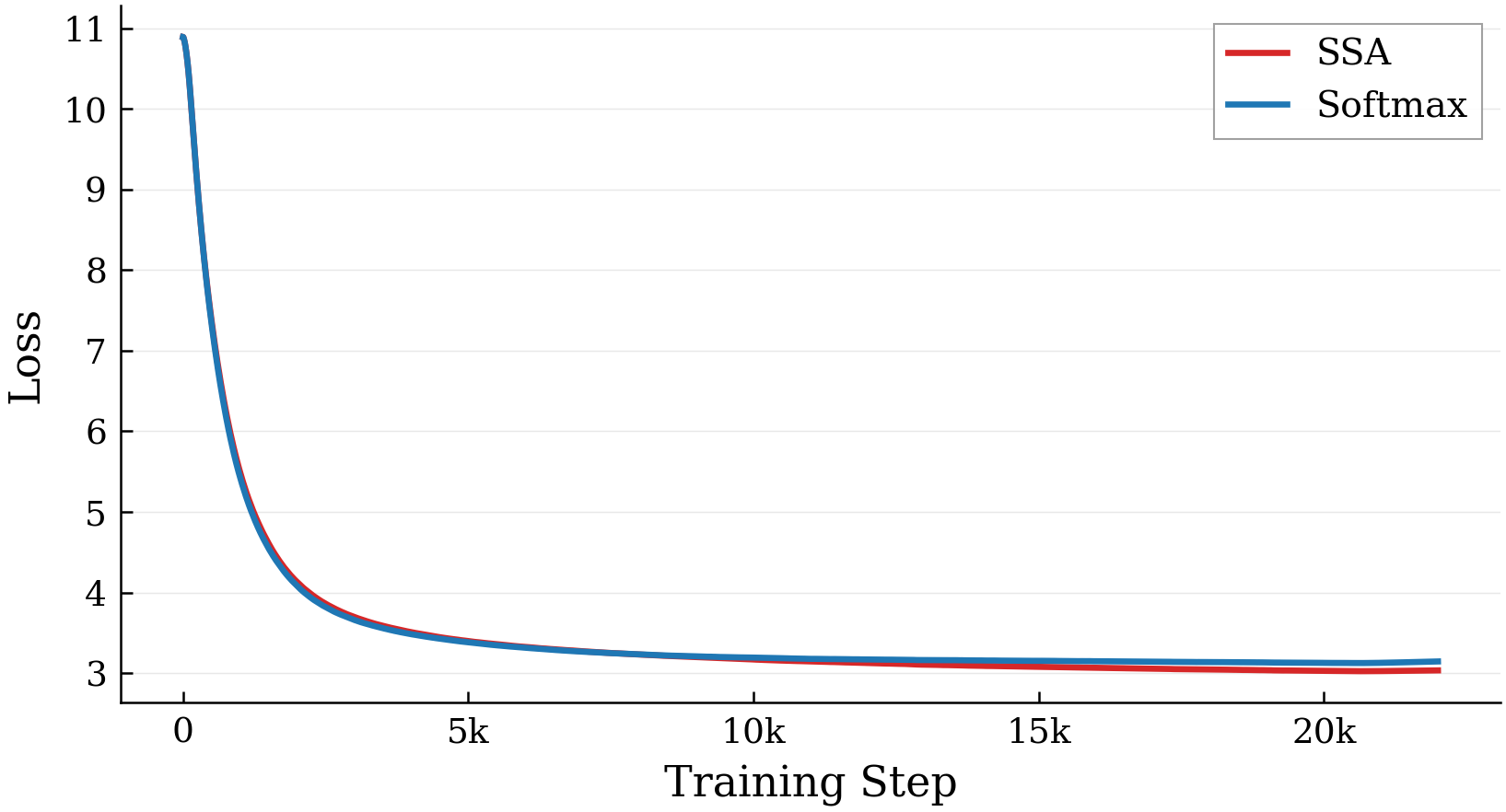}
    \caption{Training loss curves for SSA and Softmax on the 
    Nemotron-style decoder model (114M) trained on FineWeb for 22K 
    steps.}
    \label{fig:loss_curves}
\end{figure}

%\newpage
\section{Repartition of token norms in the OpenWebText}
\label{appendix:token}
\begin{figure}[!ht]
\center 
\includegraphics[width=\columnwidth]{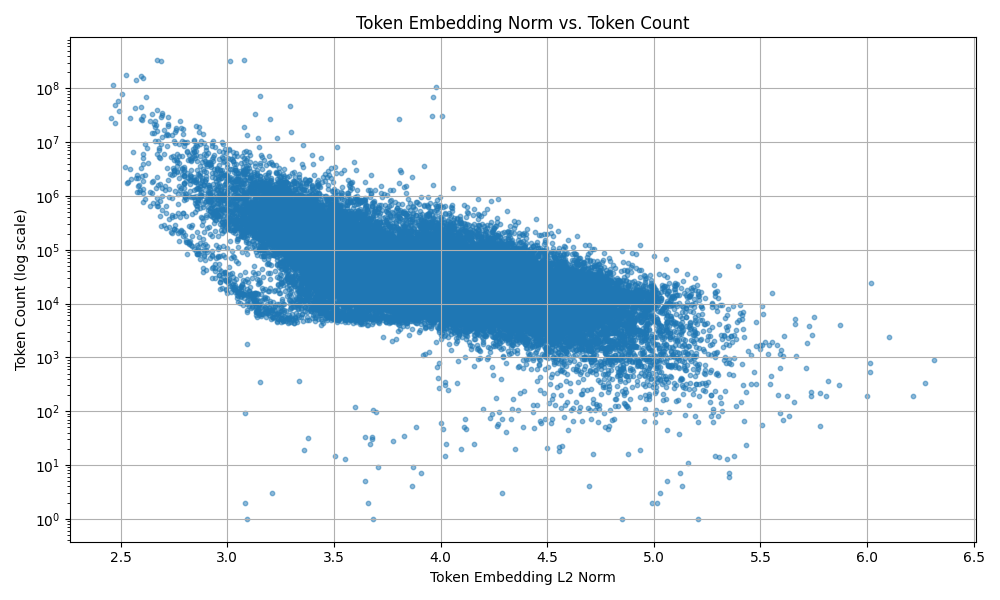}
\caption{Plot showing how many tokens have a norm of value x, for OpenWebText using GPT-2 Tokenizer. The x-axis is token embedding norm and the y-axis is token count (log scale)} 
\label{token}
\end{figure}

\hidden{
\begin{figure}[ht!]
\includegraphics[width=6cm]{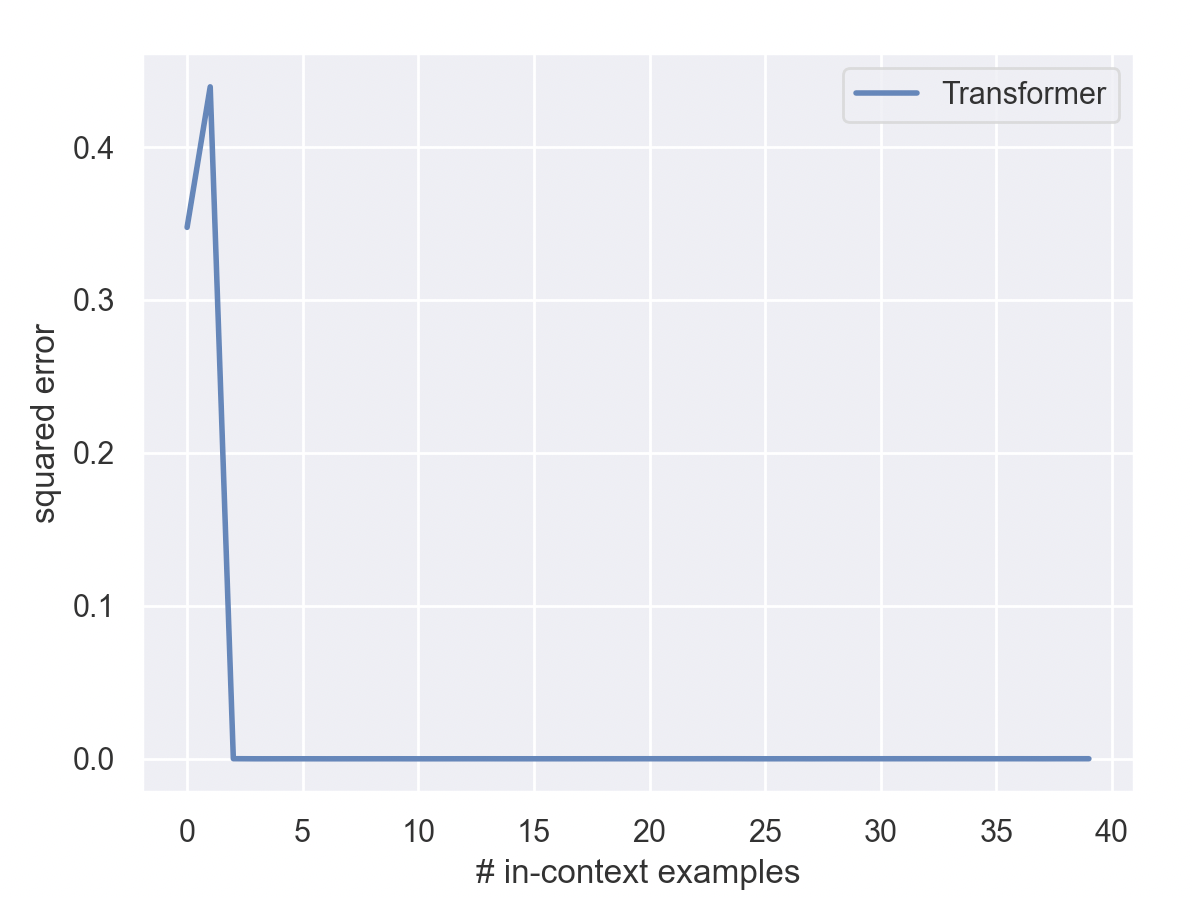} 
\includegraphics[width=6cm]{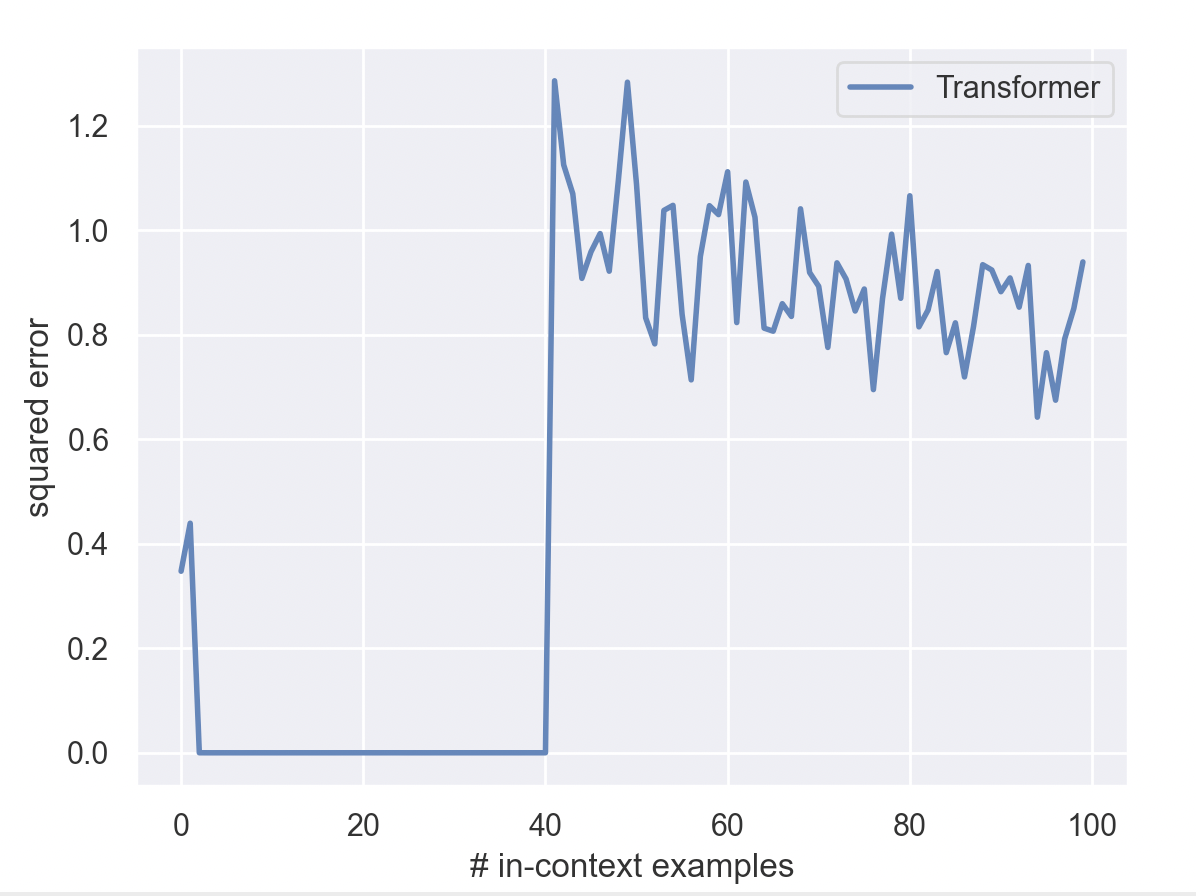}
\caption{Plot of ICL for $f(x) = x$ with $D_{\cal F}=D_{\cal I}=D_{\cal I}^t\sim U(-5,5)$ for the model 12L8AH; the one on the left is a zoom in on the first 40 points, where we see that models can often learn from 2 points, the second a view of what happens overall, when models are trained on sequences of length 41 prompts.} \label{p2>p1}
\end{figure}
}

%\newpage
\section{Heat map for the "some" task with pre-trained and finetuned models}
\label{sec:appendixOR}

\begin{figure}[!h]
\center 
\includegraphics[width=\columnwidth]{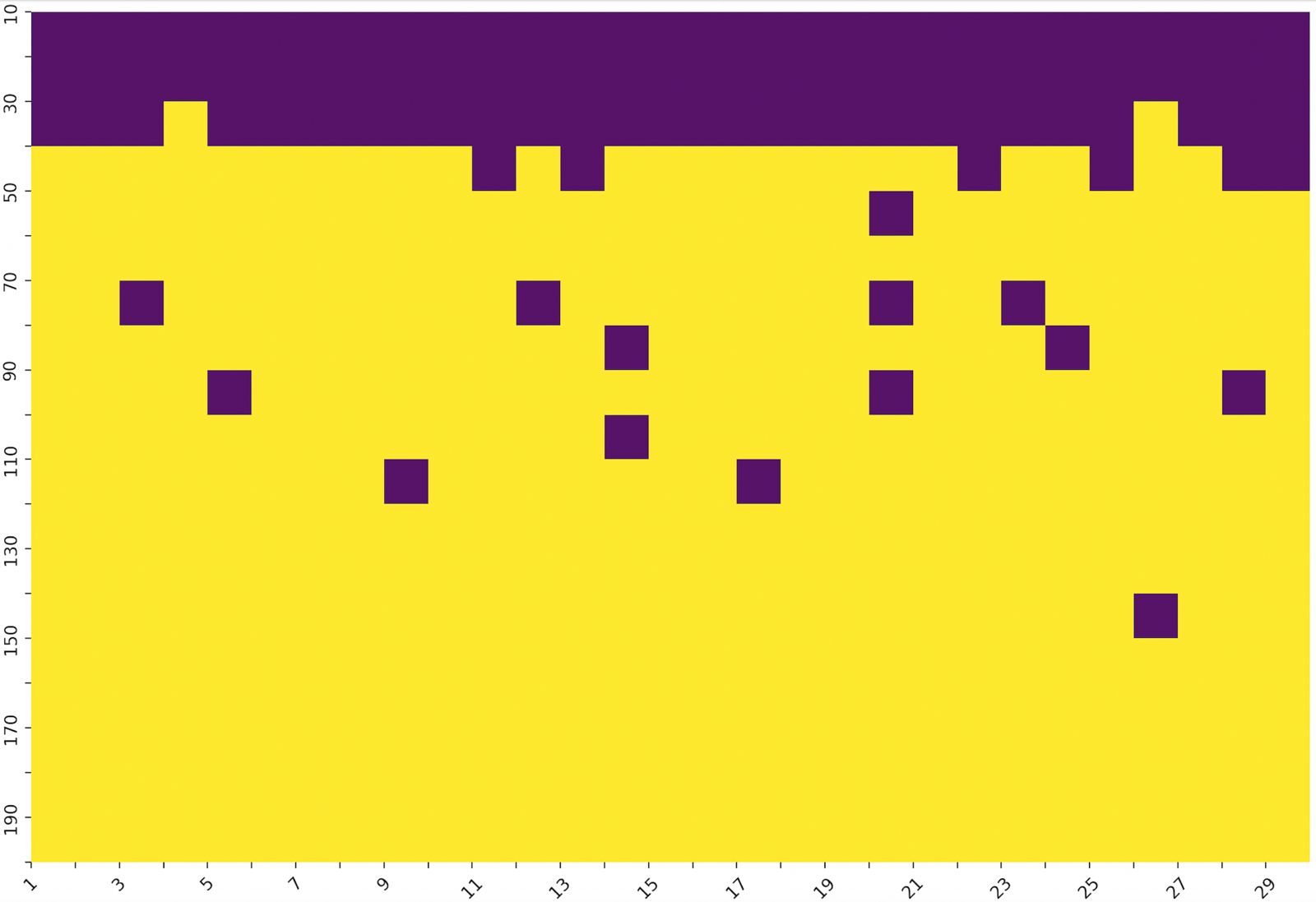}

\includegraphics[width=\columnwidth]{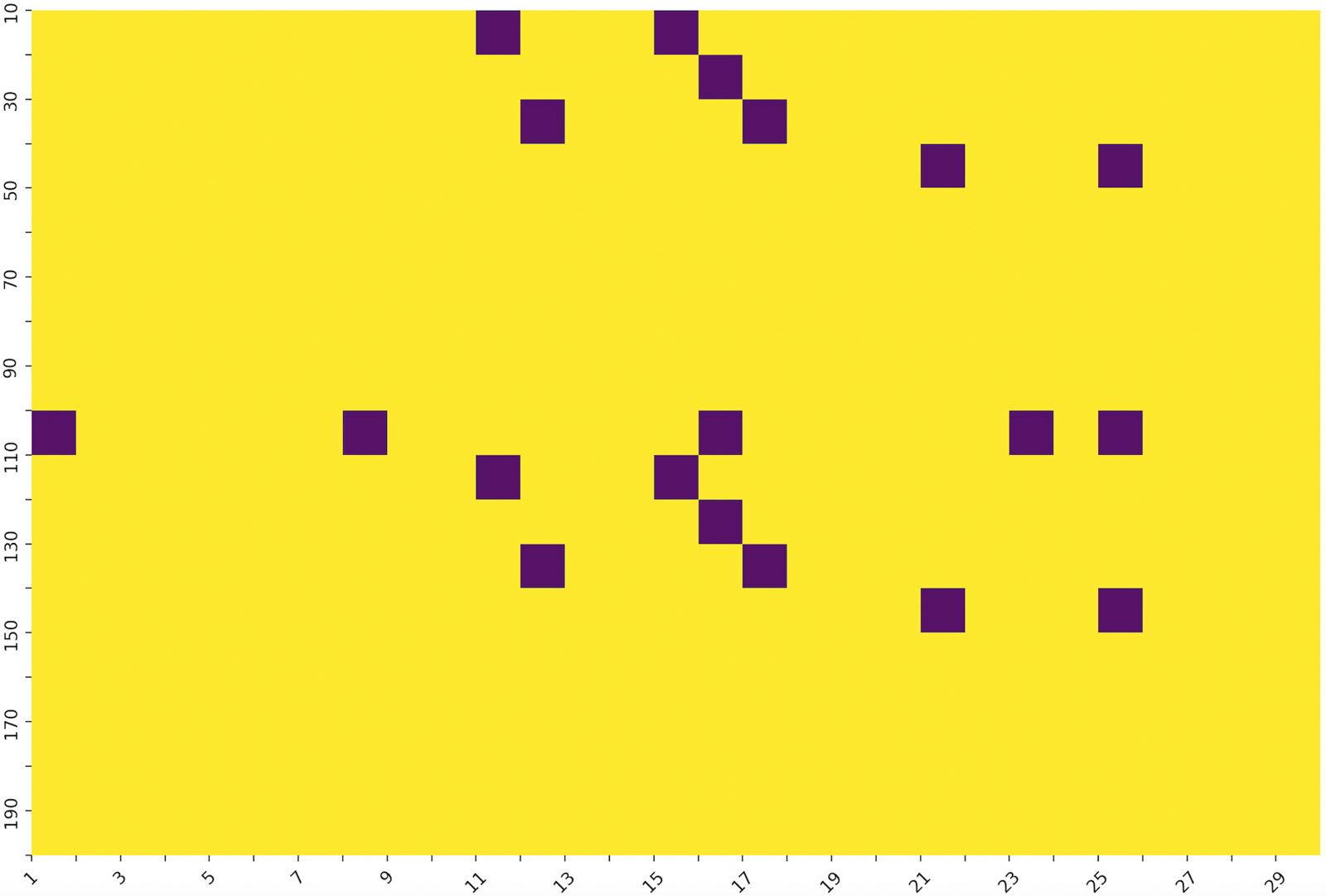}
\includegraphics[width=\columnwidth]{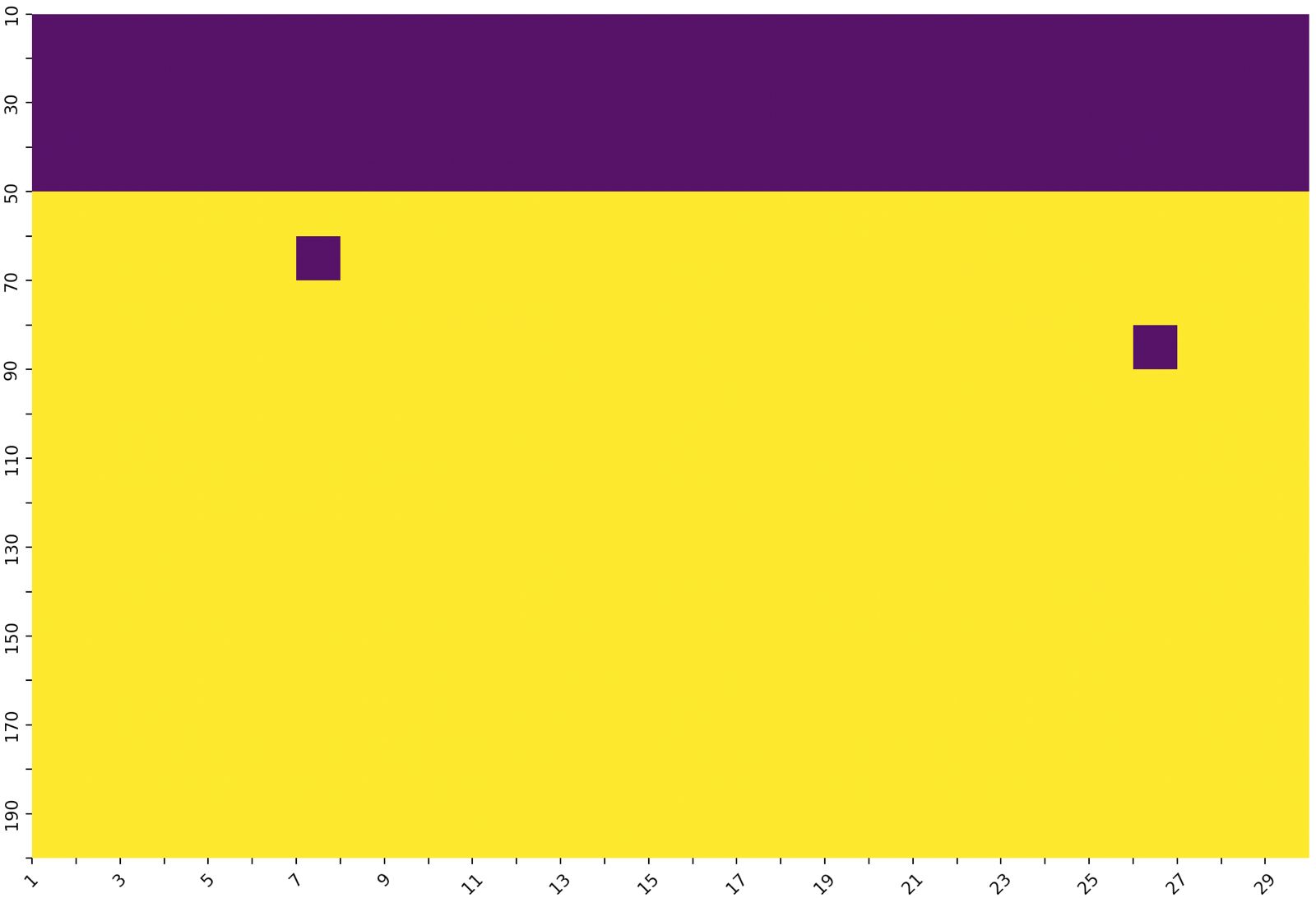}

\caption{Heatmap showing the evolution of errors for the task "some" on different models "Pre-trained Llama 3.3 70B", "Pre-trained Llama 3.1 8B" and  finetuned on Llama 31 8B", respectively from top to bottom on data in $D_{\cal I}={\mathcal N}(0,1)$ for lengths from 11 to 40 and tested in $D^t_{\cal I}={\mathcal N}(0,\sigma)$ for $\sigma \in \{1,...,10\}$ and lengths from 10 to 200. Yellow represents a much higher error rate than purple.}
\label{hmapOR}
\end{figure}
\newpage
\section{Heat map for the "every" task with pre-trained and finetuned models} 
\label{appendix:llama}

\begin{figure}[!h]

\center 

\includegraphics[width=\columnwidth]{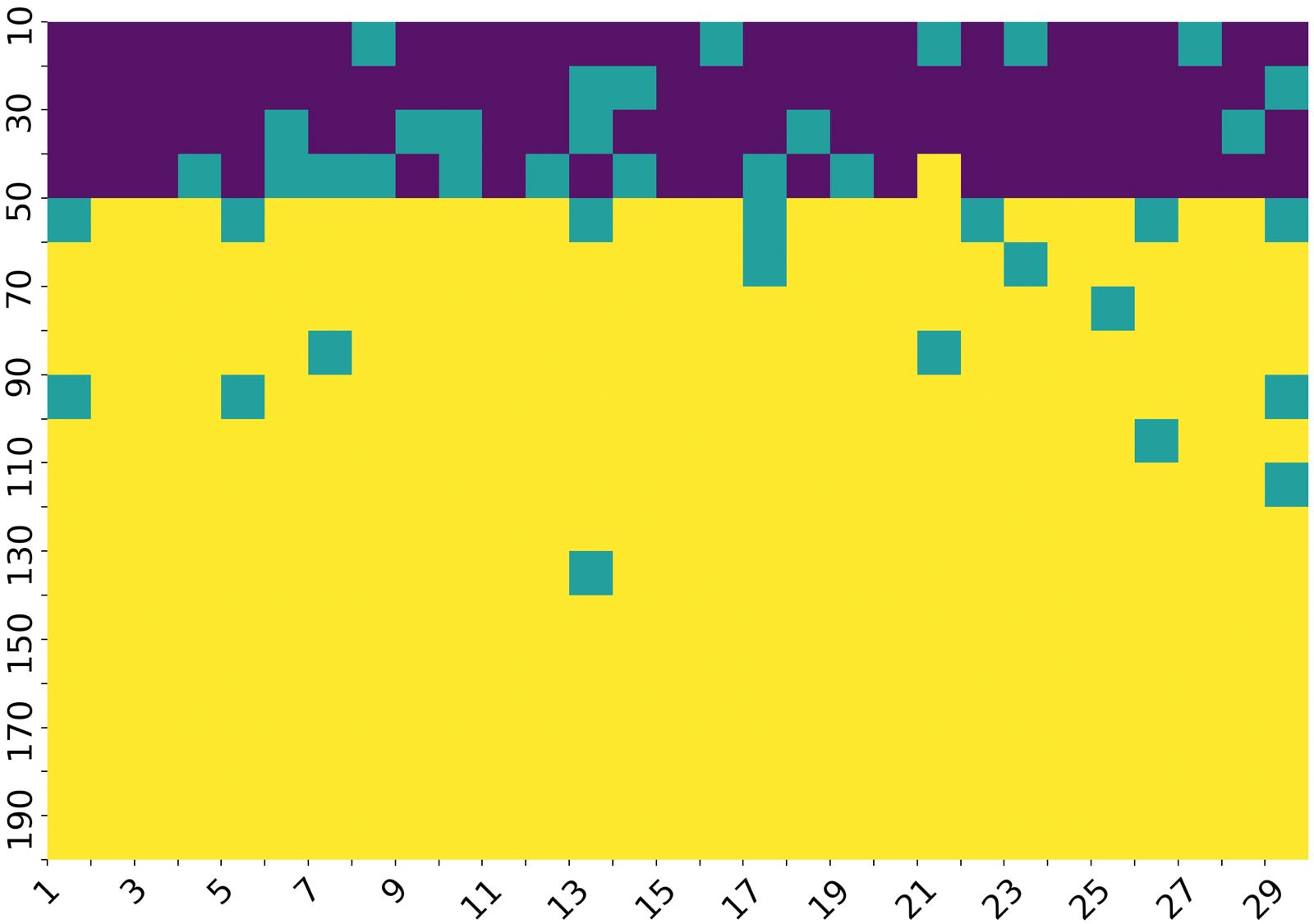}
\includegraphics[width=\columnwidth]{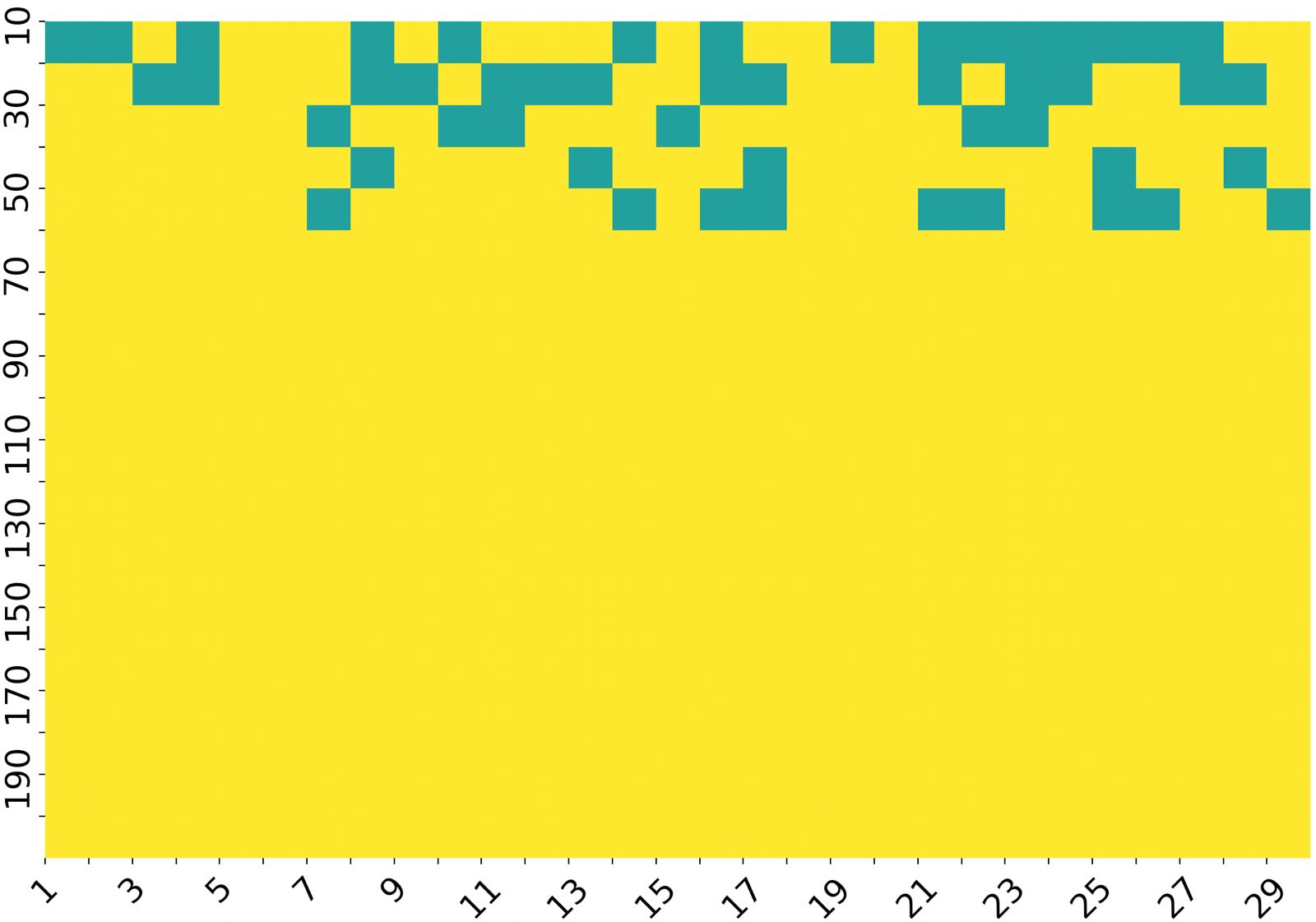}
\includegraphics[width=\columnwidth]{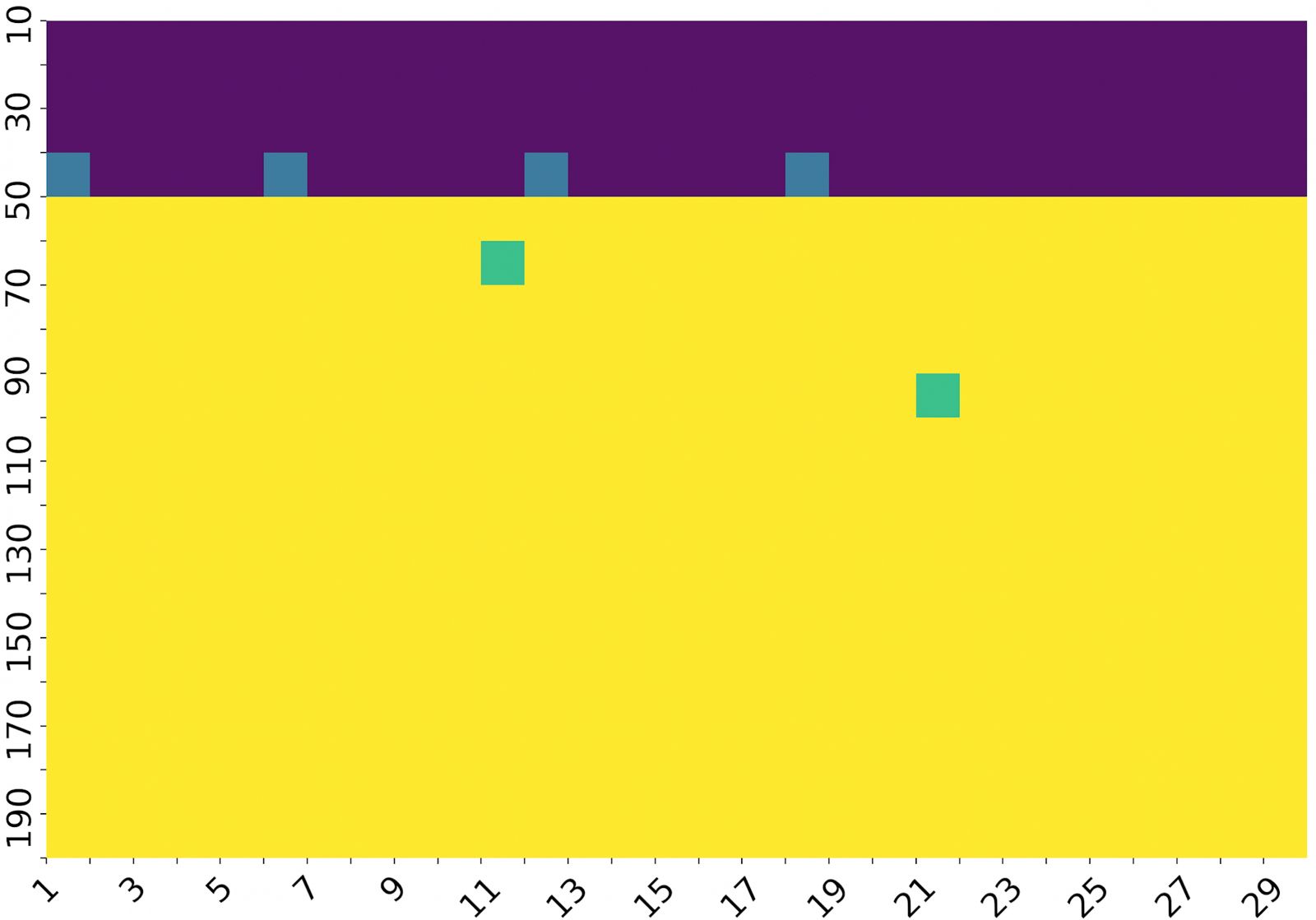}

\caption{Heatmap showing the evolution of errors for the task "every" on "Pre-trained Llama 3.3 70B" (First), "Pre-trained Llama 3.1 8B" (Second)  with 5 shot learning and  finetuned on Llama 31 8B" (Third), on data in $D_{\cal I}={\mathcal N}(0,1)$ for lengths from 1 to 40 and tested in $D^t_{\cal I}={\mathcal N}(0,\sigma)$ for $\sigma \in \{1,...,10\}$ and lengths from 10 to 200. Yellow represents a much higher error rate than purple.} 
\label{hmap3}
\end{figure}
\newpage

\section{Examples of How Models Behave In-Distribution vs. Under Distribution Shift
}
\label{appendix:bdvalues}
 \begin{figure}[!h] 
 \includegraphics[width=0.95\columnwidth]{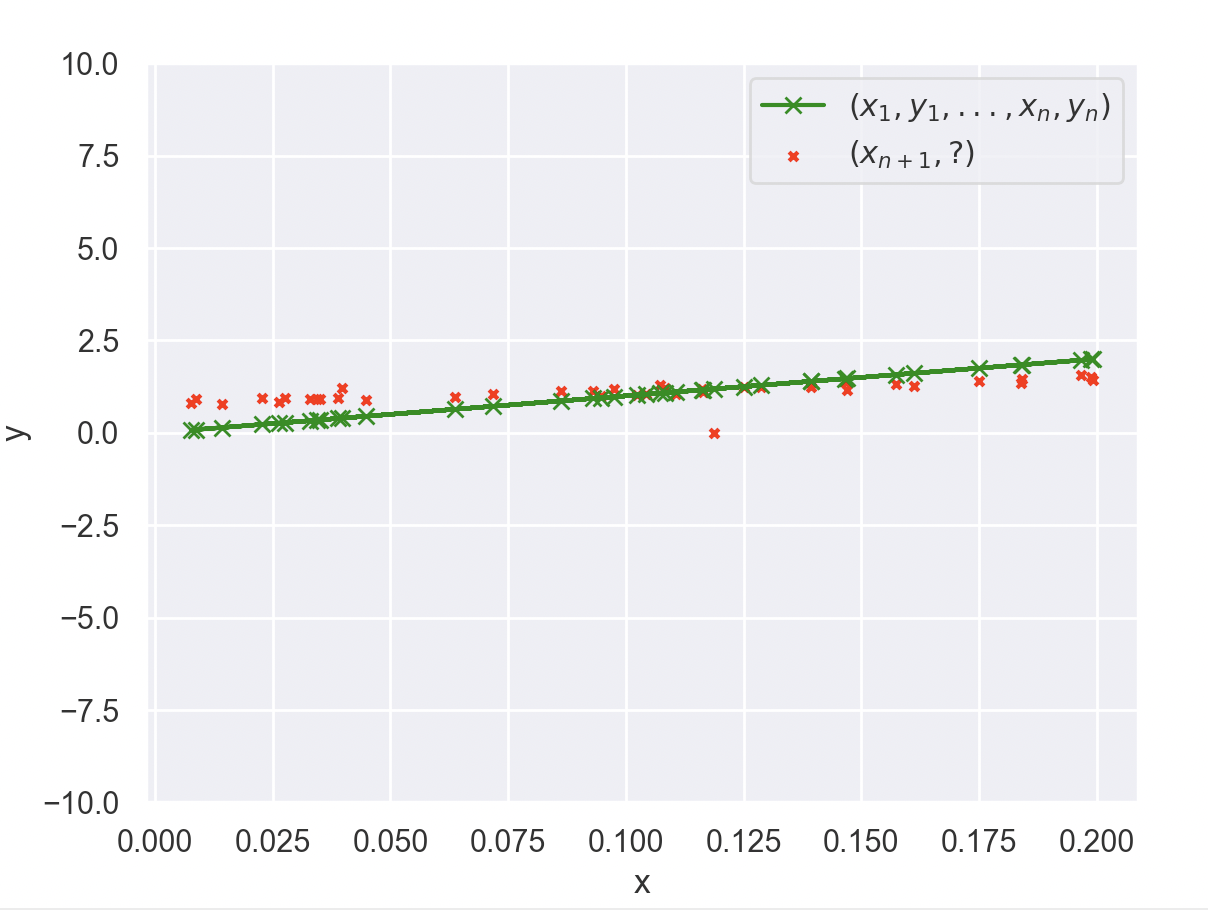}
 \includegraphics[width=\columnwidth]{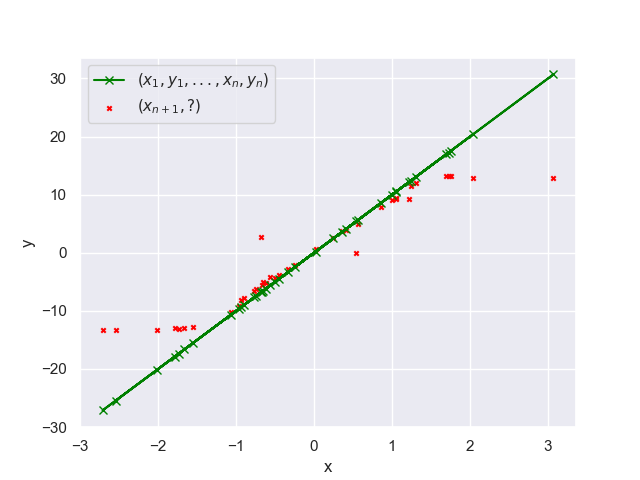} 
\includegraphics[width=\columnwidth]{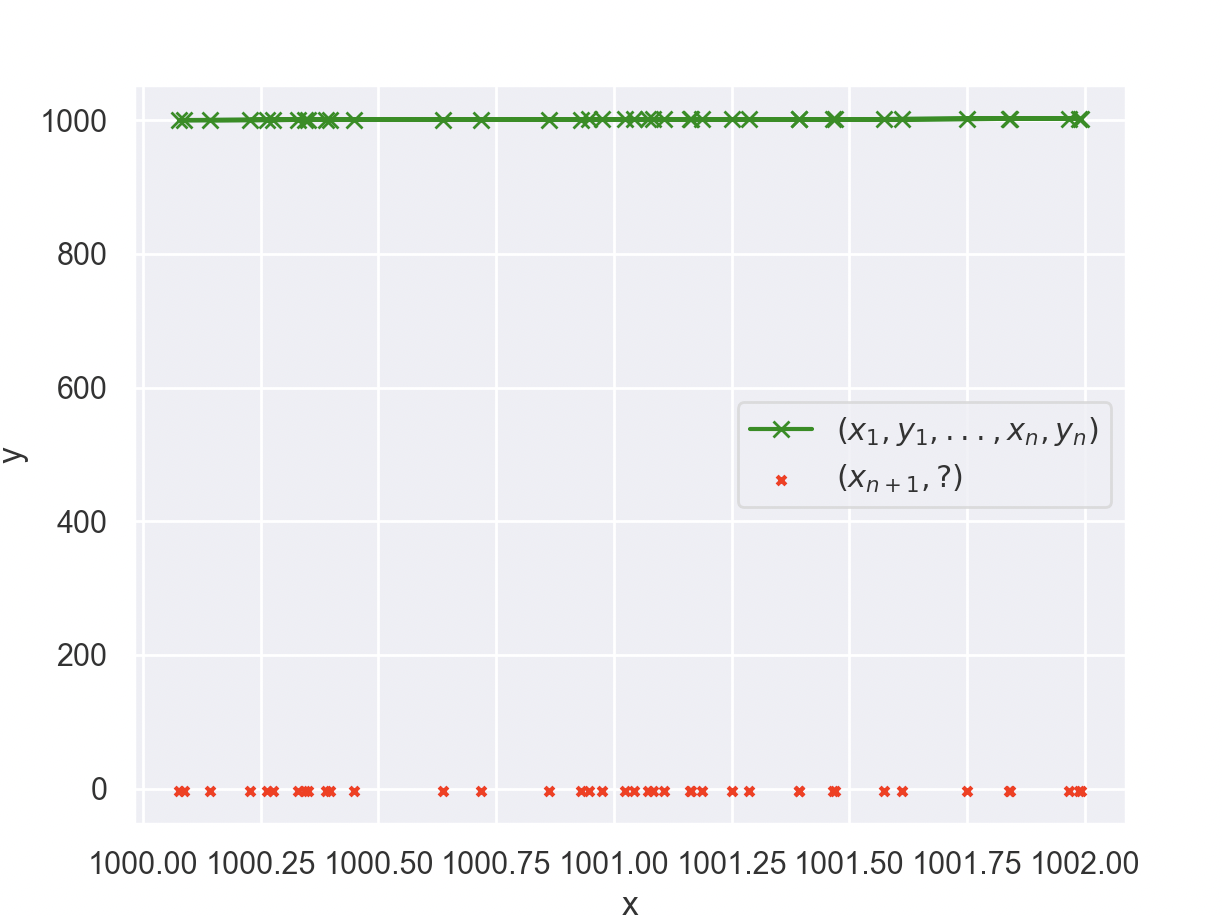}
\caption{Plots for model  12L8AH, trained on $D_{\cal I}, D_{\cal F} \sim {\mathcal N}(0,1)$  for $f(x)=x$ for high values (First) of $x$ and $f(x)=10x$ for normal (Second) then for low values of $x$ (Third)}\label{sequence}
\end{figure} 

\newpage

\section{Out-of-Distribution Generalization Failure Across Architectures}
\label{sec:appendixC}

To ensure that comparisons between models are meaningful, for each ${\mathcal N}(0,\sigma)$, we set a seed when generating the 100 random linear functions, ensuring that each model sees the same randomly chosen functions and the same set of prompting points $x_i$.    
%Figure \ref{progressive-loss} plots the evolution of error loss for various models.

\begin{figure}[!ht]
\center 
\includegraphics[width=\columnwidth]{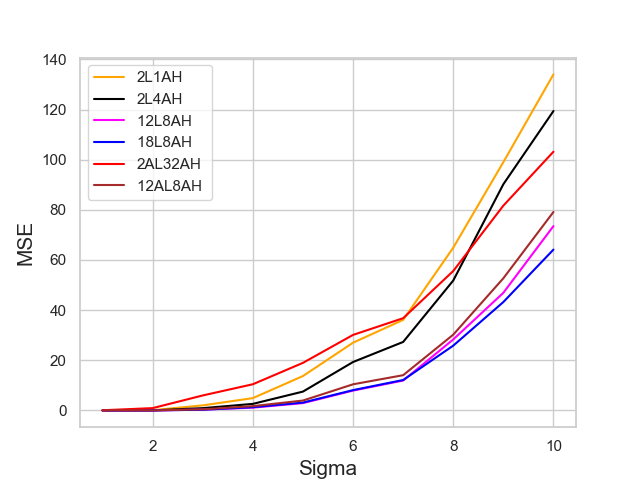}
\caption{Evolution of MSE for full transformer models (e.g., \texttt{12L8AH} = 
12 layers, 8 heads) and attention-only models (e.g., \texttt{12AL8AH} = 12 
attention layers, 8 heads), with $D_{\cal F}, D_{\cal I}, D^{test}_{\cal I} 
\sim {\mathcal N}(0,1)$ and $D^{test}_{\cal F} \sim {\mathcal N}(0, \sigma)$ 
for various $\sigma$. Both architectures exhibit strong in-distribution ICL  performance ($\sigma \approx 1$), but MSE degrades sharply for all models  as $\sigma$ increases, revealing a consistent failure to generalize 
out-of-distribution regardless of architecture.}
%And LS represents linear or ridge regression, which is trivially a perfect estimator given our totally clean input data, for attention only and full transformer models. 
\label{progressive-lossAL}
\end{figure}

\newpage

\section{Evolution of MSE over different scales of small Transformers}

\begin{table*}[ht!]
\small{
\begin{tabular}{|l|l|l|l|l|l|l|l|l|l|l|l|}
 \hline
 models \ $\backslash$ \ $\sigma$ & 1 & 10 & 20 & 30 & 40 & 50 & 60 & 70 & 80 & 90 & 100 \\ 
 \hline

  12L8AH  & 6.4e-05 & 0.05 &  0.25 & 0.50 & 0.82 & 1.21 & 1.66 & 1.87 & 2.02 & 2.14 & 2.20  \\ [1ex] 
 \hline
\end{tabular}
}
\caption{Table showing the influence of deviant inputs: Comparison showing the evolution of squared errors for models tested on $x \in D^t_{\cal I} = {\mathcal N}(0,1)$, except the first element $x_0 \in D^t_{\cal I} = {\mathcal N}(0,\sigma)$ and weights $a,b \in D^t_{\cal F}={\mathcal N}(0,1)$. }
\label{table:LF1}
\end{table*}

\newpage

\section{Prompts used for tests on pre-trained models}
\label{promptsAPP}
%The prompts are in the Figures \ref{prompt_AND}, \ref{prompt_OR} and \ref{prompt_LF}.
\label{sec:appendixF}

\begin{tcolorbox}[colback=blue!3!white,
                  colframe=blue!60!black,
                  title="Linear Function" System Prompt,
                  fonttitle=\bfseries,
                  boxrule=0.4pt,
                  arc=3pt,
                  left=6pt,
                  right=6pt,
                  top=6pt,
                  bottom=6pt]
\small
You are an auto-regressive AI model designed to predict the next value in a sequence of input-output pairs that follow a linear function $f(x) = ax + b$.
Your task is to analyze the given input-output pairs and predict the output for the final input value.

Here are some examples to illustrate the task:

\textbf{Example 1:}\\
CONTEXT: [1, 3, 2, 5, 3, 7, 4]\\
\#Answer: 9

\textbf{Example 2:}\\
CONTEXT: [0, 1, 2, 5, 4, 9, 6]\\
\#Answer: 13

\textbf{Example 3:}\\
CONTEXT: [1, -1, 2, -3, 3, -5, 4]\\
\#Answer: -7

\textbf{Example 4:}\\
CONTEXT: [0, 0, 2, 6, 4, 12, 6]\\
\#Answer: 18

\textbf{Example 5:}\\
CONTEXT: [-2, -3, 0, 1, 2, 5, 4]\\
\#Answer: 9

Now, given a new sequence of input-output pairs where the last output is missing, predict the final value.

\textbf{IMPORTANT:}
\begin{itemize}
    \item DO NOT include any explanations in your response.
    \item DO NOT use any PYTHON code in your response.
    \item GIVE JUST THE NUMERICAL OUTPUT AS THE ANSWER.
\end{itemize}
\end{tcolorbox}

\begin{tcolorbox}[colback=blue!3!white,
                  colframe=blue!60!black,
                  title="AND" Task System Prompt,
                  fonttitle=\bfseries,
                  boxrule=0.4pt,
                  arc=3pt,
                  left=6pt,
                  right=6pt,
                  top=6pt,
                  bottom=6pt]
\small
You are an auto-regressive AI model designed to evaluate whether each sublist of a list of numbers is entirely positive.
Your task is to process the list incrementally, verifying the positivity of each sublist one by one.
A sublist is considered ``TRUE'' if all its elements are greater than zero.
Once a sublist contains a non-positive number, all subsequent sublists will be marked as ``FALSE''.
Although you process the list step-by-step, you will only output the final list of booleans once all sublists have been evaluated.

Here are some examples to illustrate the task:

\textbf{Example 1:}\\
CONTEXT: [1, 1, 2, 3, -1, 2, 1]\\
\#Answer: [True, True, True, True, False, False, False]

\textbf{Example 2:}\\
CONTEXT: [0.1, -9, -0.11, 5, 0, 3.5]\\
\#Answer: [True, False, False, False, False, False]

\textbf{Example 3:}\\
CONTEXT: [-1, -2, -3, -4, -5]\\
\#Answer: [False, False, False, False, False]

\textbf{Example 4:}\\
CONTEXT: [0.5, 1.5, -0.5, 2.5, -2.5]\\
\#Answer: [True, True, False, False, False]

\textbf{Example 5:}\\
CONTEXT: [10, -10, 0, 0.01, -0.01]\\
\#Answer: [True, False, False, False, False]

Now, given a new list of numbers, perform the same task and provide the final output in the specified format.

\textbf{IMPORTANT:}
\begin{itemize}
    \item DO NOT include any other text in your response.
    \item DO NOT use any PYTHON code in your response.
    \item GIVE JUST THE OUTPUT LIST AS THE ANSWER.
\end{itemize}
\end{tcolorbox}

\begin{tcolorbox}[colback=blue!3!white,
                  colframe=blue!60!black,
                  title="OR" Task System Prompt,
                  fonttitle=\bfseries,
                  boxrule=0.4pt,
                  arc=3pt,
                  left=6pt,
                  right=6pt,
                  top=6pt,
                  bottom=6pt]
\small
You are an auto-regressive AI model designed to evaluate whether each sublist of a list of numbers has a positive element.
Your task is to process the list incrementally, verifying if there exists a positive element in each sublist one by one.
A sublist is considered ``TRUE'' if it has a positive element.
Although you process the list step-by-step, you will only output the final list of booleans once all sublists have been evaluated.

Here are some examples to illustrate the task:

\textbf{Example 1:}\\
CONTEXT: [1, 1, 2, 3, -1, 2, 1]\\
\#Answer: [True, True, True, True, True, True, True]

\textbf{Example 2:}\\
CONTEXT: [-0.1, -9, -0.11, 5, 0, 3.5]\\
\#Answer: [False, False, False, True, True, True]

\textbf{Example 3:}\\
CONTEXT: [-1, -2, -3, -4, -5]\\
\#Answer: [False, False, False, False, False]

\textbf{Example 4:}\\
CONTEXT: [-0.5, 1.5, -0.5, 2.5, -2.5]\\
\#Answer: [False, True, True, True, True]

\textbf{Example 5:}\\
CONTEXT: [-10, -10, -3, 0.01, -0.01]\\
\#Answer: [False, False, False, True, True]

Now, given a new list of numbers, perform the same task and provide the final output in the specified format.

\textbf{IMPORTANT:}
\begin{itemize}
    \item DO NOT include any other text in your response.
    \item DO NOT use any PYTHON code in your response.
    \item GIVE JUST THE OUTPUT LIST AS THE ANSWER.
\end{itemize}
\end{tcolorbox}

\hidden{
%\subsection{Prompt used for the quantifier "every"}
 \begin{figure*}[!ht] 
\includegraphics[width=15cm]{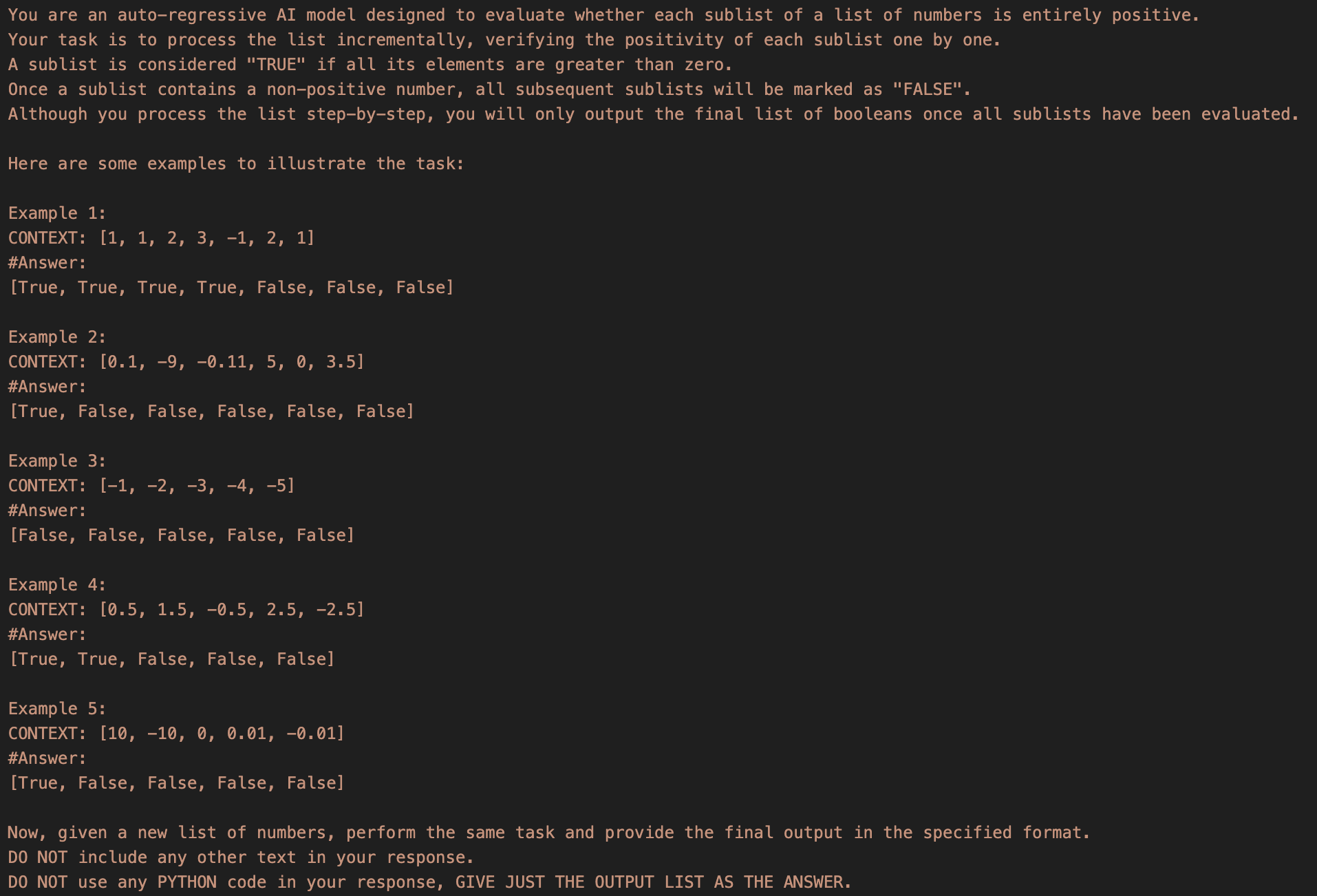}
\caption{Prompt used for the task "every"}\label{prompt_AND}
\end{figure*} 

%\newpage
%\subsection{Prompt used for the quantifier "some"}

 \begin{figure*}[!ht] 
\includegraphics[width=15cm]{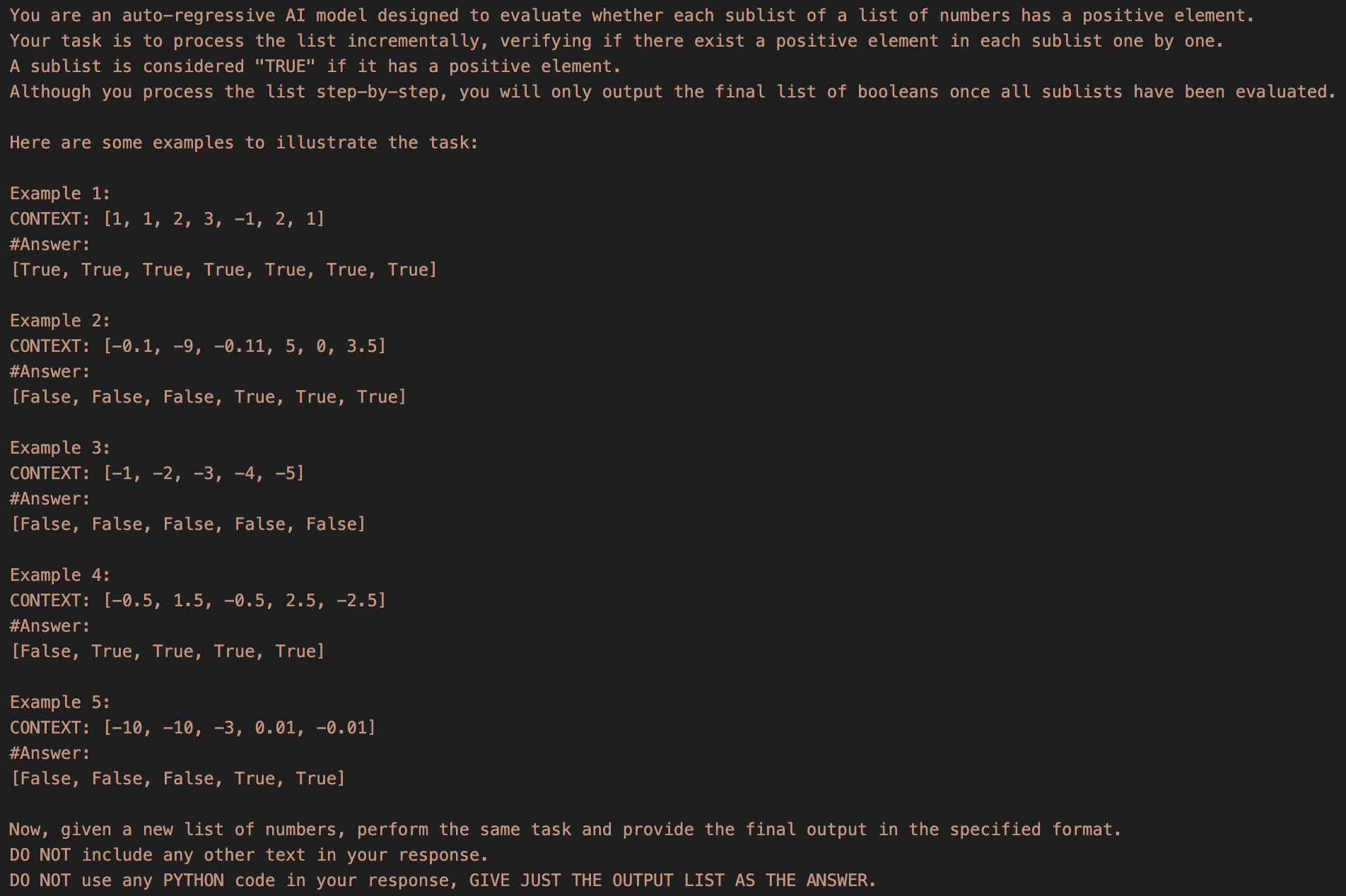}
\caption{Prompt used for the task "some"}\label{prompt_OR}
\end{figure*} 

%\newpage

%\subsection{Prompt used for linear functions}
 \begin{figure*}[!h] 
\includegraphics[width=15cm]{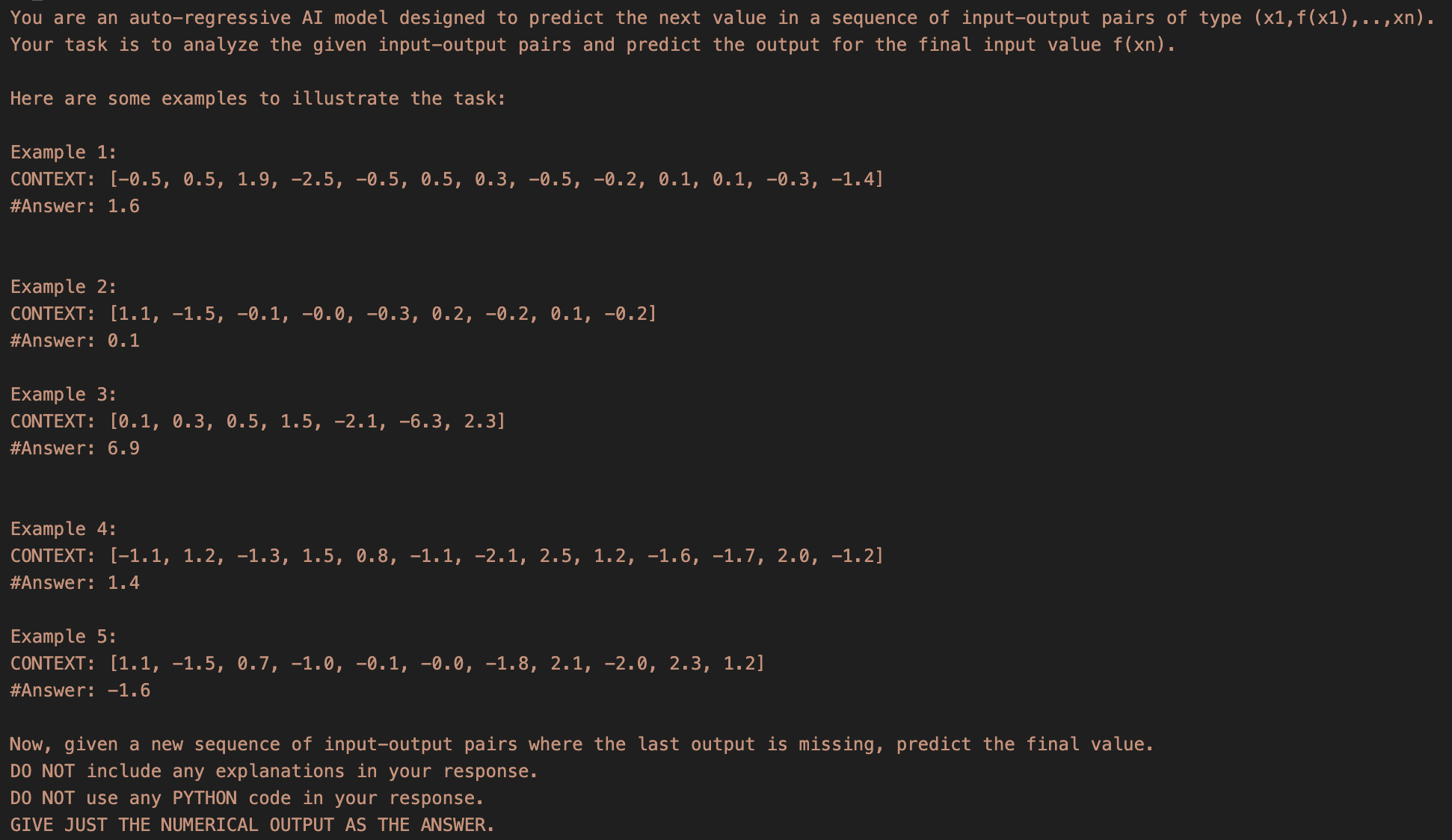}
\caption{Prompt used for the task "Linear Function"}\label{prompt_LF}
\end{figure*} 

\newpage
}

\section{Example of an output generated by a fine-tuned Llama 3.1 8B on Linear functions}
\label{appendix:lf}
Llama 3.1 8b fine-tuned on Linear Function did not understand the task and was not consistent even with the size of the output generated, which was different from an example to another. Here is an example of the generated output $[1.33,0.79, 2.61, 0.0, 0.9] $.

\end{document}